\documentclass[preprint,12pt,authoryear]{elsarticle}

\usepackage{amssymb}
\usepackage{amsmath}
\usepackage{url}
\usepackage{array}
\usepackage{booktabs}
\usepackage{dblfloatfix}
\usepackage{adjustbox}
\usepackage{multirow}
\newcommand{\myparagraph}[1]{\vspace{4pt}\noindent{\bf #1}}

\DeclareMathOperator*{\argmin}{arg\,min}

\journal{Neural Networks}

\begin{document}

\begin{frontmatter}



\title{Robust Sound-Guided Image Manipulation}


\affiliation[inst1]{organization={Department of Artificial Intelligence},
            addressline={Korea University}, 
            city={Seoul},
            postcode={02841}, 
            country={South Korea}}
\affiliation[inst2]{organization={Department of Computer Science and Engineering},
            addressline={Korea University}, 
            city={Seoul},
            postcode={02841}, 
            country={South Korea}}

\affiliation[inst3]{organization={NVIDIA Research},
            addressline={NVIDIA Corporation}, 
            city={CA},
            postcode={95051}, 
            country={USA}}

\affiliation[inst4]{organization={Graduate School of Culture Technology},
            addressline={KAIST}, 
            city={Daejeon},
            postcode={34141}, 
            country={South Korea}}

\author[inst1]{Seung Hyun Lee}
\author[inst1]{Gyeongrok Oh}
\author[inst3]{Wonmin Byeon}
\author[inst4]{Sang Ho Yoon}
\author[inst2]{Jinkyu Kim*}
\author[inst1]{Sangpil Kim*}

\begin{abstract}
Recent successes suggest that an image can be manipulated by a text prompt, e.g., a landscape scene on a sunny day is manipulated into the same scene on a rainy day driven by a text input ``raining''. These approaches often utilize a StyleCLIP-based image generator, which leverages multi-modal (text and image) embedding space. However, we observe that such text inputs are often bottlenecked in providing and synthesizing rich semantic cues, e.g., differentiating heavy rain from rain with thunderstorms. To address this issue, we advocate leveraging an additional modality, sound, which has notable advantages in image manipulation as it can convey more diverse semantic cues (vivid emotions or dynamic expressions of the natural world) than texts. In this paper, we propose a novel approach that first extends the image-text joint embedding space with sound and applies a direct latent optimization method to manipulate a given image based on audio input, e.g., the sound of rain. Our extensive experiments show that our sound-guided image manipulation approach produces semantically and visually more plausible manipulation results than the state-of-the-art text and sound-guided image manipulation methods, which are further confirmed by our human evaluations. Our downstream task evaluations also show that our learned image-text-sound joint embedding space effectively encodes sound inputs.
\end{abstract}


    
    


\end{frontmatter}
\vspace{-2em}
\let\thefootnote\relax\footnote[0]{\scriptsize{$^*$Corresponding authors:
Sangpil Kim (spk7@korea.ac.kr)} and Jinkyu Kim (jinkyukim@korea.ac.kr)}

\section{Introduction}
\label{sec:introduction}

\begin{figure}[t!]
\centering
\includegraphics[width=\linewidth]{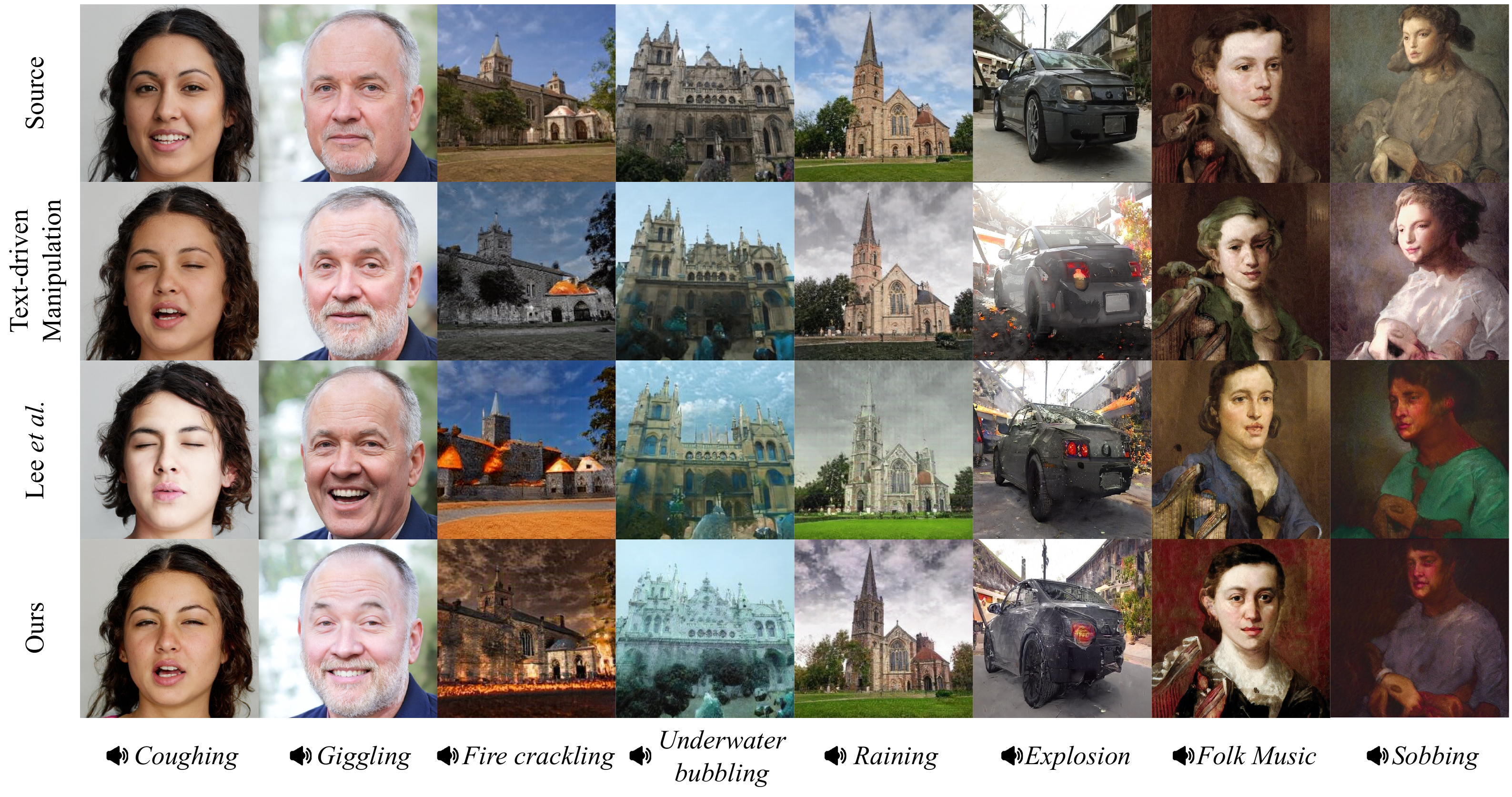}
\caption
{
Our model manipulates the given source image (top row) guided by sound inputs, e.g., coughing. We compare our image manipulation results (bottom row) with existing approaches: text-driven modified image~\citep{Patashnik_2021_ICCV} and our prior work (Lee~\textit{et al.}~\citep{lee2022sound}).
}\label{fig:tpami8}
\end{figure}

Image manipulation is a task that naturally edits images according to the user's intention. Recent studies provide guidance on manipulation using multi-modal information such as sketches~\citep{park2019semantic}, text~\citep{xia2021tedigan,Patashnik_2021_ICCV,gatys2016neural}, and sound~\citep{lee2022sound}. To apply the user's intention into the image, a mixture of sketches and text is used to perform image manipulation and synthesis~\citep{park2019semantic,xia2021tedigan}. The user's intention can be applied by drawing a picture~\citep{park2019semantic} or writing text with semantic meanings~\citep{xia2021tedigan,gatys2016neural}. We mainly focus on image modification based on user-provided sound. Because discrete input like text cannot fully express the user's intention when it is used alone, sound provides an alternative option, as it provides other properties such as continuity and dynamics over time. Such properties of sound allow image manipulation to be more dynamic and create different levels of semantics or sentiment in a scene. For example, every video clip of ``thunder'' generates a different loudness and different characteristics of ``\textit{sound of thunder}''. In addition, when a piece of music is expressed with a combination of several instruments, the atmosphere of the music can be expressed only with sound.

\begin{figure}[t!]
  \centering
  \includegraphics[width=\linewidth]{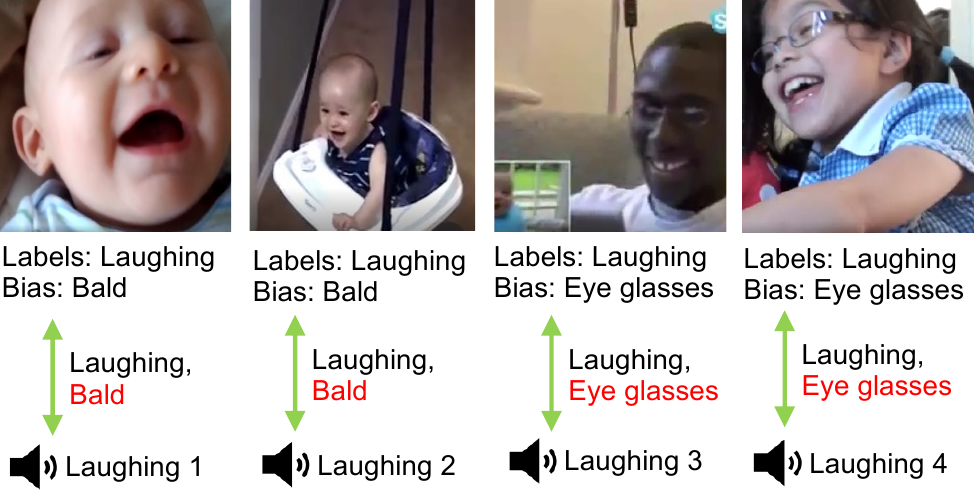}
  \caption{
  Examples of audio-visual representational biases observed in the VGG-Sound dataset~\citep{chen2020vggsound}, e.g., a scene of ``laughing'' often concurrently observed with baldness or eyeglasses.}
  \label{fig:bias}
\end{figure}

Recent text-conditional image manipulation methodologies show promising results~\citep{el2019tell, jiang2021language, li2020manigan, nam2018tagan,xia2021tedigan}. Those works use a powerful joint multi-modal embedding space which is trained on a large scale image-text pairs. These text-based prior works provide a powerful tool for image manipulation.
In particular, StyleCLIP~\citep{Patashnik_2021_ICCV}, one of the best text-guided image manipulation method, considered leveraging the representational power of Contrastive Language-Image Pre-training (CLIP)~\citep{radford2learning} models to produce text-relevant manipulations with given text input.
StyleCLIP also maintains high-quality image generation ability using StyleGAN~\citep{jeong2021tr} while allowing the style insertion of text semantics into the image. Both $\mathcal{W}$ space and extended latent space $\mathcal{W}+$, the intermediate latent space of StyleGAN, are suitable for manipulation because they disentangle the feature in the representation spaces.

Previous sound-to-visual synthesis works~\citep{chen2017deep, hao2018cmcgan, oh2019speech2face, qiu2018image, wan2019towards, zhu2021deep} have attempted to convert sound signals into realistic visual content. However, these works have a limitation in creating or changing intuitive image styles with various audio domains, as they do not leverage pre-trained joint multi-modal representations like CLIP~\citep{radford2learning}.
To solve this problem, many recent studies try to match the sound embedding space with the CLIP embedding space~\citep{wu2021wav2clip,lee2022sound,guzhov2021audioclip}.
In particular, Lee~\textit{et al.}~\citep{lee2022sound} proposed a method to provide sound-based guidance by matching the sound embedding space with the CLIP embedding space. After matching the multimodal embedding space, the StyleGAN source latent vector moves under CLIP-based guidance by minimizing the cosine distance of the manipulated image embedding and sound embedding. 
This approach allows the StyleGAN source latent vector in $\mathcal{W}+$ space to obtain a robust direction for semantic cues while being intensity-aware for various sound inputs.

\begin{figure}[t!]
  \centering
  \includegraphics[width=\linewidth]{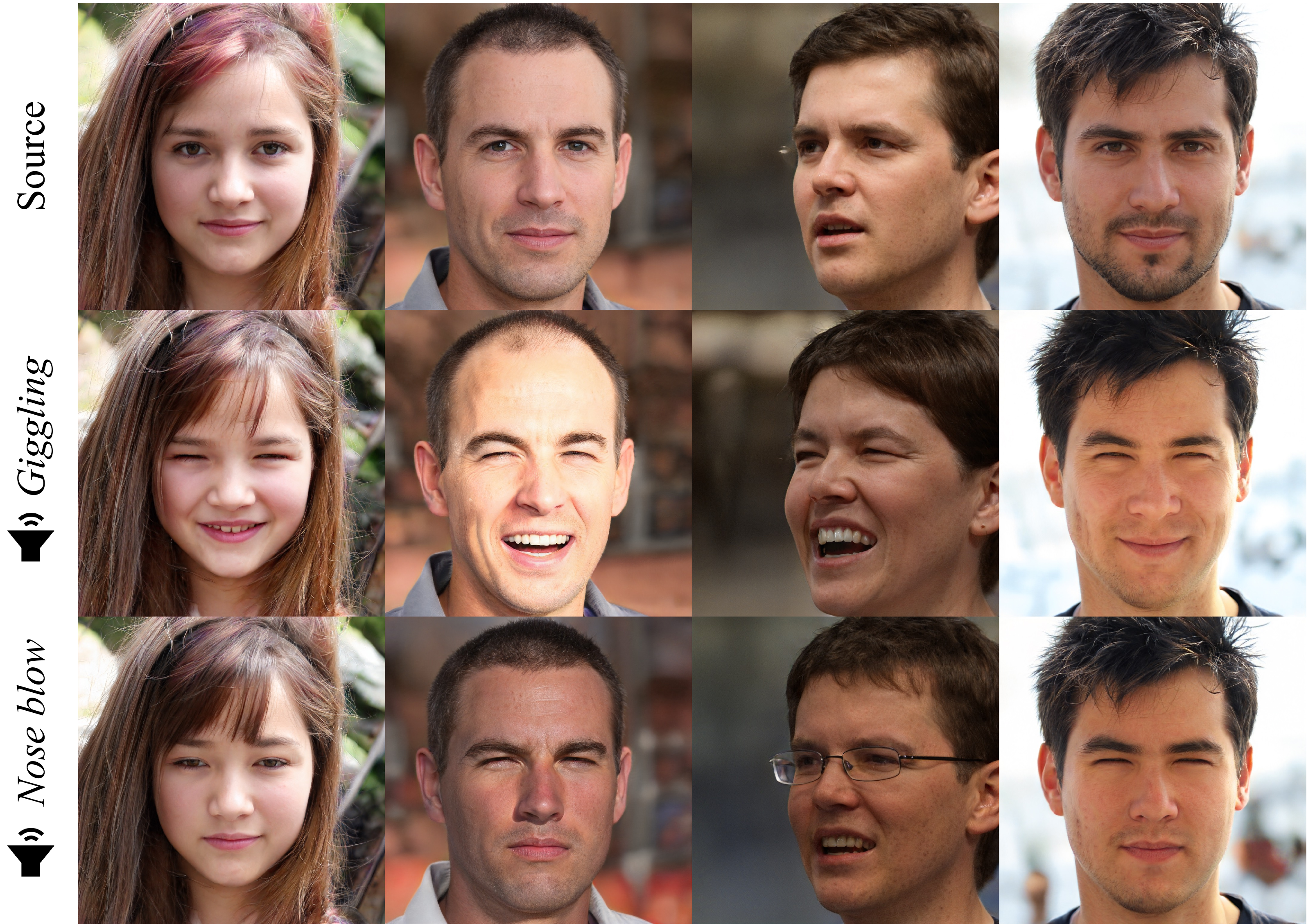}
  \caption{Failure examples of our prior sound-based image manipulation approach~\citep{lee2022sound} with sound inputs of giggling and nose blowing. Though it produces visually plausible manipulation results, fine-grained visual details (e.g., hair color) are not properly retained in the manipulated images. For example, see unintentional hair color changes (1st column), hairstyle changes (2nd column), the appearance of glasses (3rd column), and disappearance of beard (4th column).}
  \label{fig:cvprex}
\end{figure}

\begin{figure*}[t]
  \begin{center}
    \includegraphics[width=\textwidth]{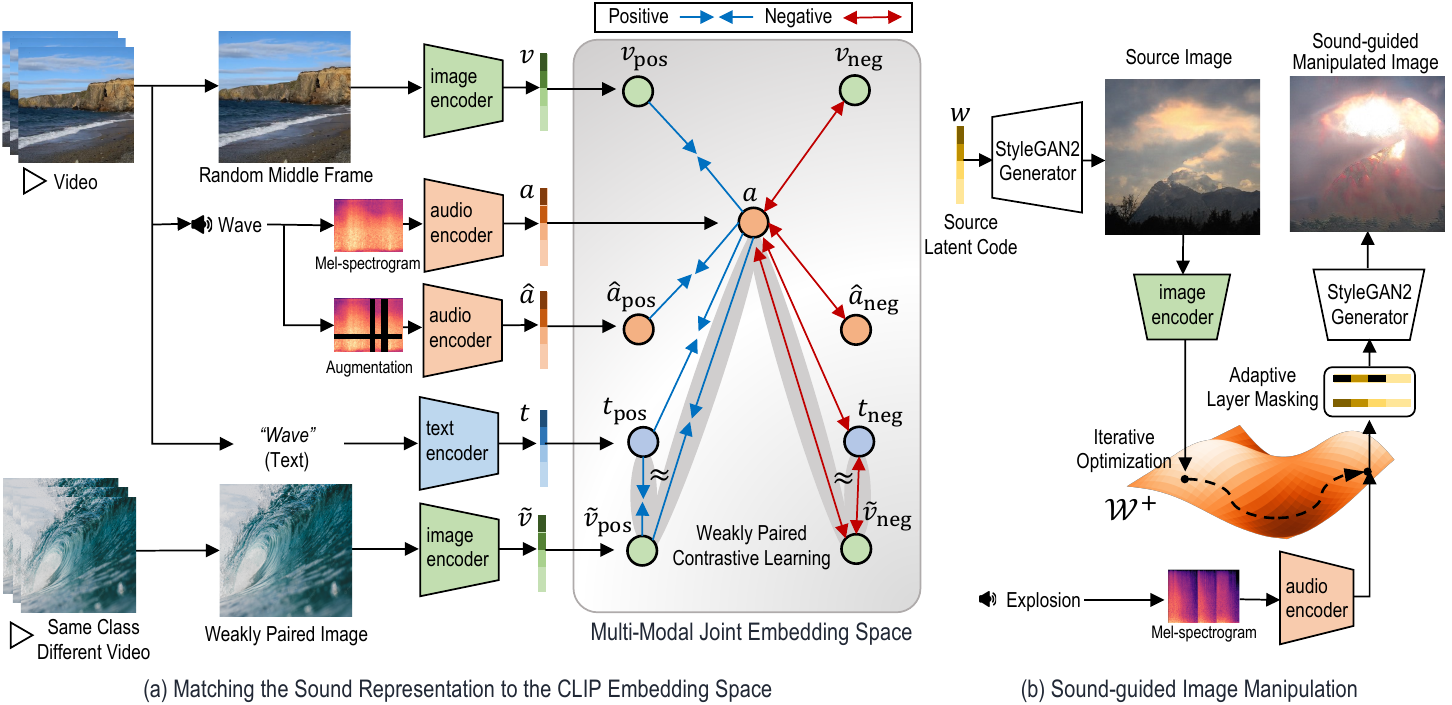}
  \end{center}
  \caption{Overview of our proposed sound-guided image manipulation process. Our model mainly consists of two steps: (a) matching the sound representation to the CLIP embedding space and (b) the StyleGAN2-based sound-guided image manipulation step. In (a), we train an audio encoder to produce the matched latent representations from a given video. The latent representations for a positive triplet pair (e.g., audio input: “Wave”, text: “wave”, and corresponding image) are mapped close together, while that of negative pair samples are mapped further away in the (CLIP-based) embedding space~(see right). To further enhance the representational power of the joint embedding space, we use a weakly-aligned image-audio pair, i.e., given a text input ``Wave'', we randomly sample an image of the same class but from other video clips. We observe that our proposed Weakly Paired Contrastive Learning effectively retains fine-grained visual details, addressing issues raised in Fig.~\ref{fig:cvprex}.  In (b), we use a direct code optimization approach where a source latent code is modified in response to user-provided audio, producing a sound-guided image manipulation result (right).
  }
  \label{fig:overview}
\end{figure*}

However, the existing sound-guided image manipulation technique~\citep{lee2022sound} has a weakness when guiding unintentional information that is not included in the sound inputs. As the sound latent vector does not project correctly into the image embedding space, the image attributes unrelated to the audio are changed. Fig.~\ref{fig:cvprex} shows how a sound sample gives audio-irrelevant direction such as eyeglasses or hair color. 
This is because the previous method~\citep{lee2022sound} allows one video sample to have only one strong positive edge between its own audio and frame even if many same class video samples share common properties. 
Specifically, we find some biases in audio-visual pairs for sound representation learning which result in fatal issue about image manipulation~(see Fig.~\ref{fig:bias}). 
As a result, sampling only one strong positive pairs from one video affects the image manipulation results as shown in Fig.~\ref{fig:cvprex}.

To tackle these problems, we propose a novel audio-visual weakly paired contrastive learning, which prevents sound representation from learning biases between audio-visual from other videos in minibatches. Our approach enlarges the number of positive weak edges between sound and image which are sampled from other videos. 
By leveraging weakly connected audio-visual pairs which shares same text description, we take advantage of the fact that videos having the same labels share common visual feature information.

We propose to minimize Kullback–Leibler divergence~\citep{hinton2015distilling} between visual-text cosine similarity probability distribution and audio-visual cosine similarity probability distribution.
Our weakly paired contrastive learning proceeds as follows. First, we sample additional visual data with text queries during the sound representation pre-training step. After that, our audio-visual similarity score aims to imitate the visual-text similarity score. 
Instead of estimating the identity matrix directly, the proposed method equalizes the audio-visual softmax-probability distribution and the visual-text softmax-probability distribution in minibatches. We find that our approach generates more robust images to fine-grained visual details, such as drastic color and hairstyle changes, than the previous method~\citep{lee2022sound}.

While expanding the scope and scale of the audio-visual relationship, we adopt~\citep{lee2022sound}'s sound representation learning process to express the relationship between audio-text and audio-audio as shown in  Fig~\ref{fig:overview}~(a). The audio encoder is trained to produce a latent representation aligned with textual and visual semantics by leveraging the representation power of pre-trained CLIP models. 
We perform self-supervised learning between audio samples to represent the inherent properties of audio. Self-supervised approach provides different views even in the same class, enabling richer sound intensity-aware image manipulation. 

After we pre-train the multi-modal joint embedding space, we use the direct latent code optimization to produce a semantically meaningful image in response to a user-provided sound as shown in Fig~\ref{fig:overview}~(b). StyleGAN's source latent vector moves in StyleGAN latent space through an iterative process that minimizes the distance between input audio and the generated image in CLIP space to generate an image that matches the meaning of the audio.

Our experimental results illustrate that the proposed method shows a robust image manipulation performance for the target attribute~(see Fig.~\ref{fig:tpami8}). Furthermore, the sound-based approach supports more diverse and detailed image manipulation results related to audio intensity information than state-of-the-art text-based image manipulation methods. In addition, we demonstrate that our sound representation outperforms existing audio-visual joint representation~\citep{wu2021wav2clip,guzhov2021audioclip,lee2022sound} for image manipulation.
We provide a wide range of image manipulation scenarios for real-world applications, such as music style transfer and sound-guided artistic painting manipulation. 
This manuscript extends the preliminary conference version of this work~\citep{lee2022sound}. In this present study, our main contributions are listed as follows: 
\begin{itemize}
    \item We propose a novel method for robust sound-guided image manipulation. We obtain a more robust sound representation with audio-visual weakly paired contrastive learning. Furthermore, our approach reduces the fine-grained bias of sound-guided image manipulation results.
    \item We demonstrate that our method is effective for sound-guided image manipulation solely based on given audio features, including temporal context, tone, and volume. In particular, the visual change of style is closely proportional to the audio intensity.
    \item We conduct much more investigations and incorporate discussions, such as ablation studies of another sound representation for image manipulation and music style transfer, and we include discussions. We consider the method's present limitations and potential future directions.
\end{itemize}

\section{Related Work}
\subsection{Text-Guided Image Manipulation}
Text-guided image manipulation approaches have been most commonly explored in the computer vision community. Several studies utilized a GAN-based encoder-decoder architecture to preserve the visual features in the manipulation process driven by descriptive text inputs~\citep{dong2017semantic, li2020manigan, nam2018tagan}. Recently, StyleCLIP~\citep{Patashnik_2021_ICCV} utilized the latent space of the pre-trained StyleGAN and a CLIP~\citep{radford2learning}-based text-image joint embedding space to synthesize visually more plausible images~\citep{Patashnik_2021_ICCV}. Further, TediGAN~\citep{xia2021tedigan} also explored using a GAN inversion technique based on multi-modal mapping. These approaches generate manipulated images by minimizing the CLIP distance between an input text and an image generated by a StyleGAN-based generator. Similarly, in this work, we also follow this strategy to optimize distances in the CLIP embedding space, but we extend the CLIP space into an image-text-sound joint embedding space, improving its representational power for the image manipulation task. We empirically observe that a sound can convey a more complex semantic context in a scene (e.g., temporal context, music genre, tone, and volume), producing a more visually plausible image manipulation.

\begin{figure*}
    \begin{center}
        \includegraphics[width=\textwidth]{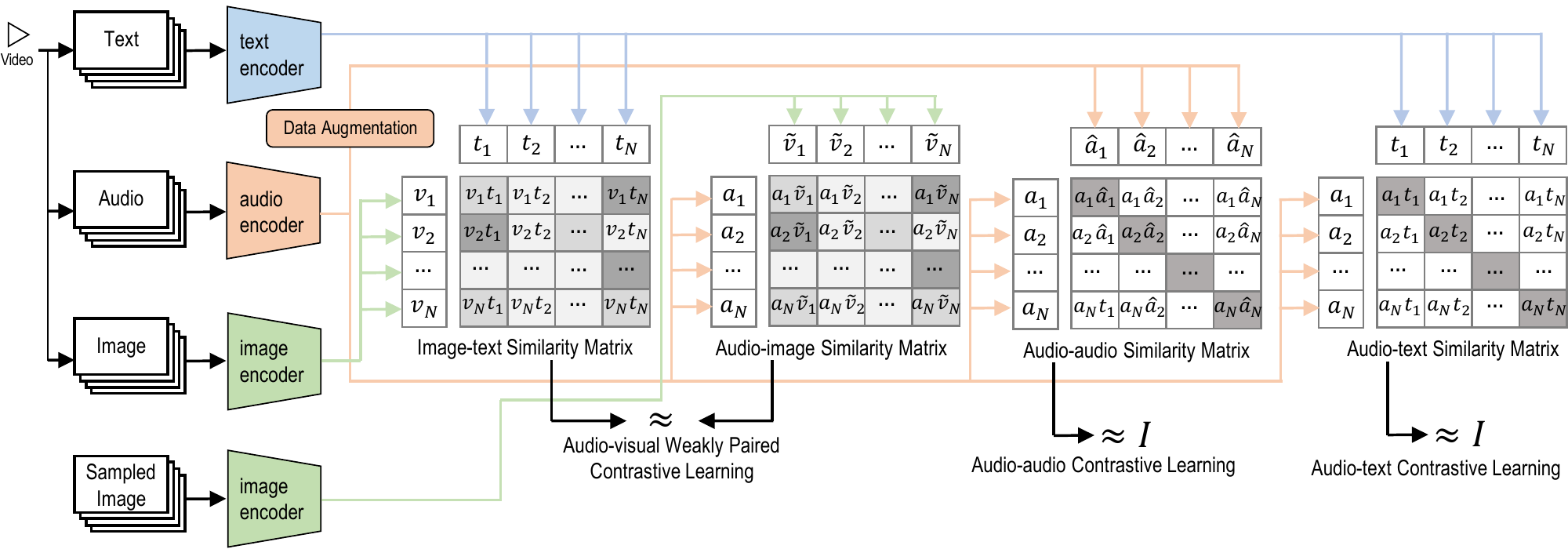}
    \end{center}
    \caption{Sound representation learning for image manipulation. We adopt different strategies for representation due to the characteristics of each modality. 
    First, we employ audio-visual weakly paired contrastive learning to obtain similarity scores between images and audio. Instead of performing direct contrastive learning between audio and image embedding, we learn that the audio-visual similarity matrix is equal to the text-visual similarity matrix.
    Second, we use intensity-aware self-supervised learning to separate different scenes in the embedding space, even if the audio class is the same. Finally, in order to have a joint embedding space between audio and text, we perform contrastive learning between the audio and text embeddings.}
    \label{fig:contrastivelearning}
\end{figure*}

\subsection{Sound-Guided Image Manipulation}
A few works have explored the use of sound inputs in the image manipulation task. However, to our best knowledge, most of these works focused on music instead of sound semantics. Lee~\textit{et al.}~\citep{lee2020crossing} explored a music-to-visual style transfer model with a cross-modal learning strategy. Jeong~\textit{et al.}~\citep{jeong2021tr} introduced a neural music visualizer that maps the latent representation of given music to visual embeddings of StyleGAN. Similarly, \textit{Tr$\ddot{a}$umerAI}~\citep{jeong2021tr} used the latent transfer mapping of music-to-StyleGAN's embedding to visually express a music input. However, they do not focus on sound semantics (but only on the reaction) in the navigational direction in the StyleGAN latent space, limiting the representational advantages of using sound. For example, \textit{Crossing you in Style}~\citep{lee2020crossing} used the era of music to define the semantic relationship between sound and images, and thus it can only transfer to a limited number of image styles.

In our prior work~\citep{lee2022sound}, we studied a novel image manipulation approach with sound input. We first demonstrated that a CLIP-based image-text-sound joint embedding space could be learned via a contrastive learning technique. We also reported that such a joint embedding space can be used to manipulate images driven by various sound inputs (environmental sound, music, etc.). However, this model often fails to retain semantic visual cues that the user-provided sound should not manipulate. This work addresses this issue using audio-visual weakly paired contrastive learning and intensity-aware self-supervised learning techniques.

\subsection{Interpreting Latent Space in StyleGAN}
The intermediate latent space in pre-trained StyleGAN~\citep{karras2019style} solves the disentanglement issue and allows the generated images to be manipulated meaningfully according to changes in the latent space. 
Extended latent space $\mathcal{W}+$ allows image manipulation with interpretable controls by a pre-trained GAN generator~\citep{abdal2019image2stylegan, karras2019style, karras2020analyzing}. 
For latent space analysis in audio sequences, \textit{Audio-reactive StyleGAN}~\citep{brouwer2020audio} generates an image at every time step by calculating the magnitude of the audio signal and moving it in the latent space of StyleGAN. 

However, the method cannot control the meaning of sound in the latent space. The magnitude of the sound only determines the movement of the latent vector in the StyleGAN latent space. We can manipulate images with the properties of sound using the pre-trained sound representation. Instead of randomly traversing in StyleGAN latent space, our model can give reasonable sound semantics guidance.

\subsection{Audio-Visual Representation Learning}
Audio-visual representation learning aims to obtain the generalized relationship between different modalities to utilize in several audio-visual downstream tasks, such as audio classification. Most audio-visual representation learning studies~\citep{DBLP:journals/corr/AytarVT17,nagrani2018learnable, suris2018cross}  are designed to map distinct modalities into the same embedding space. The correlation between modalities is learned by audio-visual contrastive learning~\citep{chen2021distilling, mazumder2021avgzslnet, sun2020learning}. Contrastive learning allows the positive audio-visual pairs to have more similar representations than the representations of other negative pairs.

However, audio-visual representation learning is still challenging for image manipulation because there is as much positive pairs as CLIP~\citep{radford2learning} to learn the correlation between different modalities.
CLIP learned the relationship between image and text embedding through the multi-modal self-supervised learning of 400 million image-text pairs and show a zero-shot inference performance comparable to supervised learning in most image-text benchmark datasets.

To overcome this problem, recent studies~\citep{wu2021wav2clip,guzhov2021audioclip,lee2022sound} align the sound embedding space with the CLIP embedding space. In order to predict the same CLIP image representations from the audio stream, Wav2CLIP~\citep{wu2021wav2clip} freezes the CLIP image encoder and trains the audio encoder on the audio streams from videos. AudioCLIP~\citep{guzhov2021audioclip} extends modality and scale by using a newly collected trimodal audio-visual-text dataset. Lee~\textit{et al.}~\citep{lee2022sound} trains an audio encoder from scratch with a audio-text dataset and audio-visual streams from video to obtain a CLIP-based sound latent representation. 

In this work, we additionally sample images that are suitable for text queries, and we propose a novel method of weakly paired contrastive learning to form a link between two modalities in sound representation~\citep{lee2022sound}. Given triplets, we sample other images from text descriptions and learn audio-visual weak supervision. In this way, the bias of the audio-visual representation is reduced, which produces a much more realistic sound-guided image manipulation result.

\section{Method}
In this work, we present a novel method for sound-guided image manipulation. Our model consists of the following two main steps: (i) the CLIP-based multi-modal latent representation learning~(Section~\ref{section:clip}) and (ii) the sound-guided image manipulation~(Section~\ref{section:manipulation}). As we will explain in Section~\ref{section:clip}, we utilize the CLIP-based image-text joint embedding space and extend it to deal with an additional modality, i.e., sound. However, learning such a joint representation from scratch is generally challenging due to the lack of multi-modal datasets. Thus, we instead leverage the pre-trained CLIP model, which optimized a visual-textual joint representation by using contrastive learning. Then, we train our audio encoder to produce representations that align with other modalities. We achieve this by applying contrastive learning losses for audio-text and audio-audio pairs. Additionally, we improve the quality of joint representations by applying a novel weakly paired contrastive loss for audio-image pairs.

In Section~\ref{section:manipulation}, we explain the sound-guided image manipulation step. Given the multi-modal joint embedding space, we use a direct latent code optimization method to manipulate images guided by an audio input. For example, a facial expression of a portrait photo is manipulated according to sound inputs, e.g., giggling or sobbing, as shown in Fig.~\ref{fig:tpami8}. Note that our approach may be similar to existing text-guided image manipulation models, including StyleCLIP~\citep{Patashnik_2021_ICCV}, but we focus on exploring a novel modality, i.e., sound.

\subsection{Multi-Modal Latent Representation Learning} \label{section:clip}
\subsubsection{Extending CLIP Embedding Space with Sound}\label{section:text}
As shown in Fig.~\ref{fig:contrastivelearning}, we train a set of encoders with three different modalities \{audio, text, and image\} to produce the matched representations in the embedding space from a video dataset. 
Specifically, given audio, text, and image inputs from the same video, i.e. $x_a$, $x_t$, and $x_v$, we use three different encoders to obtain a set of $d$ dimensional latent representations, i.e. $\bf{a}$, $\bf{t}$ and ${\bf{v}}\in\mathbb{R}^{d}$, respectively.

To learn these joint latent representations, we use a typical contrastive learning approach that follows the work by Radford~\textit{et al.}~\citep{radford2learning}: the latent representations for a positive triplet pair are mapped close together in the embedding space while representations of negative pair samples are mapped farther away from each other. Specifically, we use InfoNCE loss~\citep{alayrac2020self} to map positive audio-text pairs and audio-visual pairs close together in the CLIP-based joint embedding space, and we map negative pairs farther away from each other. 
Formally, given a minibatch of $N$ audio-text representation pairs $\{{\bf{a}}_i, {\bf{t}}_j\}$ for $i\in\{1, 2, \dots, N\}$, we first compute the $(i,j)$ component of the audio-text similarity matrix $M^{a\rightarrow t} \in \mathbb{R}^{N \times N}$ between the ${\bf{a}}_i$ and ${\bf{t}}_j$ latent representation as:
\begin{equation}
    M^{a\rightarrow t}_{(i, j)}=\cfrac{\exp({\bf{a}}_i \cdot {\bf{t}}_j/\tau) }{\sum_{\textnormal{k=1}}^N\exp({\bf{a}}_i \cdot {\bf{t}}_k/\tau)},
    \label{loss:matrix}
\end{equation}
where ${\bf{a}}_i \cdot {\bf{t}}_j$ represents vector dot product and $\tau$ is a temperature hyperparameter. We emphasize that ${\bf{a}}_i$ and ${\bf{t}}_j$ are normalized vectors, so that the magnitude of each vector is 1. This loss function is the log loss of an $N$-way classifier that seeks to predict the $i$-th diagonal component $\{{\bf{a}}_i, {\bf{t}}_i\}$ as the true representation pair in the similarity matrix $M^{a\rightarrow t}$. 
\begin{equation}
    l_{i}^{(a\rightarrow t)}=-\text{log}(M^{a\rightarrow t}_{(i, i)}).
    \label{loss:mini_con}
\end{equation}
As the loss function is asymmetric, we also define the following text-to-audio similarity matrix $M^{t\rightarrow a}$:
\begin{equation}
    M^{t\rightarrow a}_{(i, j)}=\cfrac{\exp({\bf{t}}_i \cdot {\bf{a}}_j/\tau) }{\sum_{\textnormal{k=1}}^N\exp({\bf{t}}_i \cdot {\bf{a}}_k/\tau)}.
    \label{loss:matrix2}
\end{equation}
We maximize the diagonal component of the matrix $M^{t\rightarrow a}$ by minimizing the InfoNCE loss as:
\begin{equation}
    l_{i}^{(t\rightarrow a)}=-\text{log}(M^{t\rightarrow a}_{(i, i)}).
    \label{loss:mini_con2}
\end{equation}
Concretely, we minimize the following loss function $\mathcal{L}_{\textnormal{nce}}$ as a sum of the two losses $l_{i}^{(a\rightarrow t)}$ and $l_{i}^{(t\rightarrow a)}$ for all positive audio-text representation pairs in each minibatch of size $N$:
\begin{equation}
    \begin{aligned}
    \mathcal{L}_{\textnormal{nce}}^{(a \leftrightarrow t)}=\cfrac{1}{N}\sum_{i=1}^N (l_{i}^{(a\rightarrow t)} + l_{i}^{(t\rightarrow a)}).
    \end{aligned}
    \label{loss:nce}
\end{equation}

At the same time, we also calculate the diagonal component of the audio-visual similarity matrix $M^{a\rightarrow v}$ in the same way as in Eq.~\ref{loss:mini_con2}. We determine the following $\mathcal{L}_{\textnormal{nce}}^{(a \leftrightarrow v)}$:
\begin{equation}
    \begin{aligned}
    \mathcal{L}_{\textnormal{nce}}^{(a \leftrightarrow v)}=\cfrac{1}{N}\sum_{i=1}^N (l_{i}^{(a\rightarrow v)} + l_{i}^{(v\rightarrow a)}),
    \end{aligned}
    \label{loss:nce_visual}
\end{equation}
where $l_{i}^{(v\rightarrow a)}$ and $l_{i}^{(a\rightarrow v)}$ denote the negative log-likelihood of $M^{v\rightarrow a}_{(i, i)}$ and $M^{a\rightarrow v}_{(i, i)}$ in each minibatch of size $N$.

\myparagraph{Data Augmentations.}
We further apply a data augmentation strategy to improve the quality of representations and overcome the lack of large-scale audio-text multi-modal datasets. For the text inputs, we augment the text data by (i) replacing words with synonyms, (ii) applying a random permutation of words, and (iii) inserting random words. Note that, for (i), we find synonyms of the given word from WordNet~\citep{fellbaum2010wordnet} and insert the synonym at random into the given text input. For example, we augment the original texts {\it ``rowboat, canoe, kayak rowing"} to produce the new text {\it ``row canoe, kayak quarrel rowboat."}

\begin{figure}[t!]
  \centering
  \includegraphics[width=\linewidth]{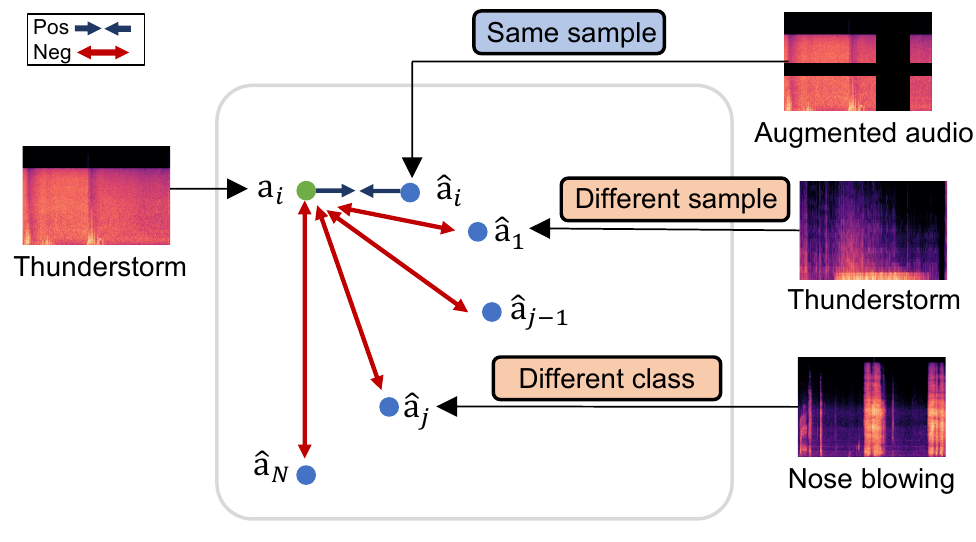}
  \caption{
  Intensity-aware self-supervised representation learning for audio inputs. Augmented audio of the same scene is considered a positive pair, and other samples of the different scenes are considered a negative pair (even if they are the same class).
  }
  \label{fig:self-supervised}
\end{figure}

\subsubsection{Self-Supervised Learning towards Intensity-Aware Sound Representation}
\label{section:audio}
Unlike text inputs, audio inputs can provide more diverse and find-grained semantic cues. For example, a "thunderstorm" can be expressed in various ways: e.g., weak or severe thunderstorms, thunderstorms with heavy rain, winds, hail, or no precipitation. It may be challenging for texts to verbalize and differentiate these subtle differences, but it does not apply to sound. Thus, we apply self-supervised learning techniques between audio samples (e.g., weak vs. severe thunderstorms) to obtain an intensity-aware sound representation. 

Existing self-supervised learning approaches depend on a contrastive loss that makes representations of the same-class different views close in the embedding space. In contrast, different-class views are pushed farther away. To this end, we regard the  audio input of different augmented views as positive and different audio samples as negative. Note that we obtain a latent representation $\hat{{\bf{a}}}\in\mathbb{R}^{d}$ from an augmented audio input $\hat{x}_a$, which has been shown to be useful for improving the quality of the latent representation, as this is a common practice in self-supervised representation learning.

Formally, we define a similiarity matrix $M^{a\rightarrow \hat{a}} \in \mathbb{R}^{N\times N}$ from embeddings obtained from audio and augmented audio. The $(i,j)$ component of the audio-audio similarity matrix $M^{a\rightarrow \hat{a}}$ is as follows:
\begin{equation}
    M^{a\rightarrow \hat{a}}_{(i, j)}=\cfrac{\exp({\bf{a}}_i \cdot {\bf{\hat{a}}}_j/\tau) }{\sum_{\textnormal{k=1}}^N\exp({\bf{a}}_i \cdot {\bf{\hat{a}}}_k/\tau)}.
    \label{loss:matrix_audio}
\end{equation}

The intensity-aware representation is obtained by minimizing the InfoNCE loss of $M^{a\rightarrow \hat{a}}_{(i, i)}$. We apply this technique to improve the quality of audio representations by minimizing $\mathcal{L}_{\textnormal{self}}^{(a \leftrightarrow \hat{a})}$ as follows:
\begin{equation}
    \mathcal{L}_{\textnormal{self}}^{(a \leftrightarrow \hat{a})}=\cfrac{1}{N}\sum_{i=1}^N (l_{i}^{(a\rightarrow \hat{a})} + l_{i}^{(\hat{a}\rightarrow a)}),
    \label{loss:self}
\end{equation}
where $l_{i}^{(a\rightarrow \hat{a})}$ and $l_{i}^{(\hat{a}\rightarrow a)}$ are defined in a similar way, as in Eq.~\ref{loss:mini_con} and \ref{loss:mini_con2}. This loss function is useful for learning subtle differences over sound inputs, as it must maximize the mutual information between two different views of the same inputs but to minimize the mutual information between two views of the different inputs. For example, as shown in Fig.~\ref{fig:self-supervised}, an audio sample $a_i$ forms a negative pair with $\hat{a}_j$ for $i\neq j$, which induces a diffusive effect in the embedding space. For audio inputs, we apply the SpecAugment~\citep{park19e_interspeech}, which visually augments Mel-spectrogram acoustic features by warping the features and masking blocks of frequency channels.

\begin{figure}[t!]
  \centering
  \includegraphics[width=\linewidth]{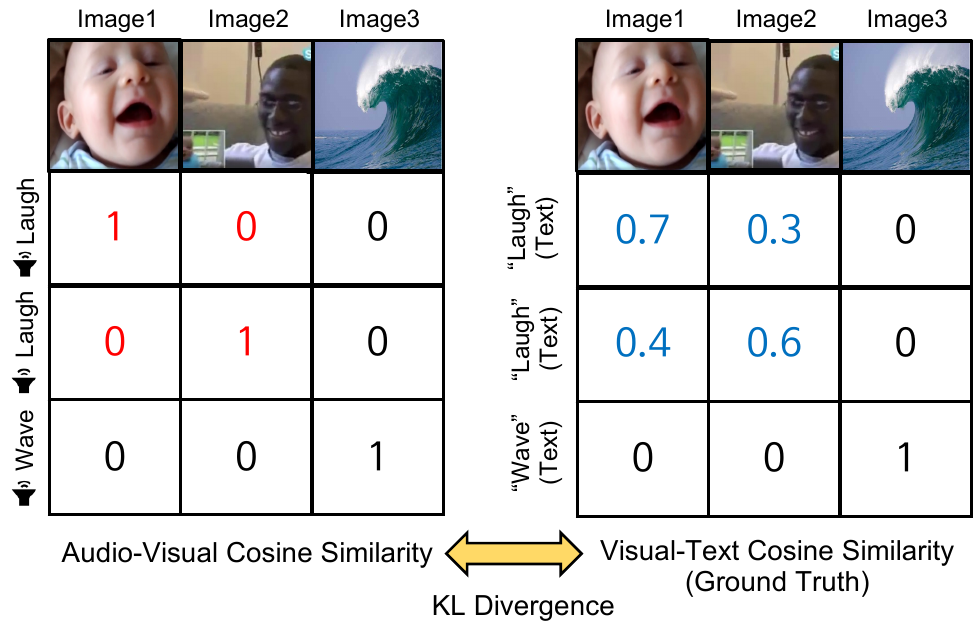}
  \caption{
  A graphical example of weakly paired contrastive learning for audio-visual joint embedding space. We minimize KL divergence between (left) audio-visual cosine similarity distribution and (right) visual-text cosine similarity distribution instead of simply enforcing it into the identity matrix, reducing the negative effect of audio-visual representational biases(see examples in Fig.~\ref{fig:bias}).
  }
  \label{fig:weak}
\end{figure}
\subsubsection{Weakly Paired Contrastive Learning for Audio-Visual Joint Embedding Space}
\label{section:image}
In addition, we propose weakly paired contrastive learning for audio-visual data to ensure that the representations of the input audio and different images with the same text labels are similar. The conventional contrastive loss only uses positive pairs which have audio and image from the same video. Therefore, audio and image from different videos are considered to be negative pairs although they may share visual features. Our new loss encourages these pairs to be closed to each other if they have the same text labels. 

We first sample the new images $\tilde{x}_v$, which have the same text labels with given audio $x_a$. Then, we use the image encoder to obtain a new sampled image embedding $\tilde{v} \in \mathbb{R}^d$. We apply the knowledge distillation technique~\citep{hinton2015distilling} to transfer the generalization ability of the CLIP to sound representation. The learning scheme allows the similarity matrix between audio-visual and visual-text to be equal in mini-batch~(see Fig.~\ref{fig:weak}).

The link knowledge between the audio and the new sampled image is transferred from CLIP to the diagonal component of a new audio-visual similarity matrix.
After sampling a new triplet by sampling sound, text and new images of size $N$ in a mini-batch, 
we define the audio-visual similarity matrix from the ${\bf{a}}_i$ and $\tilde{\bf{ v}}_j$ embeddings as:
\begin{equation}
    M^{a\rightarrow \tilde v}_{(i, j)}=\cfrac{\exp({\bf{a}}_i \cdot \tilde{ \bf{ v}}_j/\tau) }{\sum_{\textnormal{k=1}}^N\exp({\bf{a}}_i \cdot \tilde{{\bf{ v}}}_k/\tau)}, 
    \label{loss:matrix_distill}
\end{equation}
where $\tau$ is a temperature hyperparameter to refine the softmax-probability distributions, as was done in~\citep{hinton2015distilling}. Specifically, we compute the pseudo cosine similarity matrix $M^{t\rightarrow \tilde{v}}$ between the latent representations of text and those of images. The audio-visual link knowledge is transferred from the fixed CLIP to $M^{a\rightarrow \tilde v}_{(i, i)}$. Then, we match the two types of logits in the sound-image and text-image similarity matrix instead of using hard labels such as $1$ for positive pairs. We minimize the following Kullback-Leibler divergence~\citep{hinton2015distilling} for the $i$-th diagonal component:
\begin{equation}
    l_{i}^{KD}= \text{KL}(M^{a\rightarrow \tilde v}_{(i, i)}, M^{t\rightarrow \tilde v}_{(i, i)}),
    \label{loss:distillation}
\end{equation}
where $\text{KL}$ denotes the Kullback-Leiber divergence. The $\text{KL}$ can be rewritten as:
\begin{equation}
    \text{KL}(M^{a\rightarrow \tilde v}_{(i, i)}, M^{t\rightarrow \tilde v}_{(i, i)})=-M^{t\rightarrow \tilde v}_{(i, i)} \log{M^{a\rightarrow \tilde v}_{(i, i)}}
    \label{loss:kd}
\end{equation}
The KL divergence loss $\mathcal{L}_{KL}$ for $N$ batches is as follows:
\begin{equation}
    \mathcal{L}_{\text{KL}}^{(a \leftrightarrow \tilde v)}=\cfrac{1}{N}\sum_{i=1}^N l_i^{KL}.
    \label{loss:distillation_sum}
\end{equation}
To summarize, we minimize the following loss function $\mathcal{L}_\text{total}$:
\begin{equation}
    \begin{aligned}
    \mathcal{L}_{\textnormal{total}} =  \mathcal{L}_{\textnormal{nce}}^{( a \leftrightarrow t)} + \mathcal{L}_{\textnormal{nce}}^{( a \leftrightarrow v)} + \mathcal{L}_{\textnormal{self}}^{( a \leftrightarrow \hat{a})}+\mathcal{L}_{\text{KL}}^{(a \leftrightarrow \tilde v)}.
    \end{aligned}
    \label{loss:con}
\end{equation}

\subsection{Sound-Guided Image Manipulation}
\label{section:manipulation}
After learning the multi-modal joint embedding space by minimizing Eq.~\ref{loss:con}, we use a direct latent code optimization method, similar to StyleCLIP~\citep{Patashnik_2021_ICCV}, to manipulate the given image. Our method optimizes for each image instance manipulated at the inference stage. Our model minimizes the distance between a given source latent code and an audio-driven latent code in the learned joint embedding space to produce sound-guided manipulated images. Moreover, we propose an {\it Adaptive Layer Masking} technique, which adaptively manipulates the latent code. 

\subsubsection{Direct Latent Code Optimization} Given user-provided audio input $x_a$, we employ the direct latent code optimization for sound-guided image manipulation by solving the following optimization problem:
\begin{equation}
    \begin{aligned}
    \ & \underset{w_a \in \mathcal{W}+} \argmin \;{\mathcal{L}_{\textnormal{cos}}(G(w_a),x_a) +  \lambda_{\text{reg}}{\mathcal{L}_{reg}}} + \lambda_{\textnormal{ID}}{\mathcal{L}_\text{ID}}(w_a), \\ 
    \end{aligned}
    \label{loss:man}
\end{equation}
where a given source latent vector $w_s\in\mathcal{W}^{L \times D}$~(the
intermediate latent space in StyleGAN), audio-driven latent code $w_a\in\mathcal{W}^{L \times D}$ is optimized. 
$L$ denotes the number of style layers at different levels, and $D$ denotes the dimension. $\mathcal{L}_{\textnormal{reg}}$ is regularization loss. $\mathcal{L}_{\textnormal{ID}}$ and $G$ are the identity loss and StyleGAN-based generator, respectively. By minimizing the hinge loss $\mathcal{L}_\text{cos}$, we find the sound-guided latent vector $w_a$ which generates an sound-guided image in the StyleGAN latent space. $\lambda_\text{reg}$ and $\lambda_{ID}$ are hyperparameters that control the strengths of the similarity loss term and the identity loss function $\mathcal{L}_{\textnormal{ID}}$. High values of $\lambda_\text{reg}$ and $\lambda_{ID}$ lead to maintenance of the content of the source image, whereas low values do not.
The source latent code $w_s$ means the randomly generated latent code from $G$ or the latent code obtained from the existing input image through GAN inversion~\citep{richardson2021encoding, 10.1145/3450626.3459838}. 

With such an optimization scheme, the hinge loss $\mathcal{L}_{\textnormal{cos}}$ is calculated as:
 \begin{equation}
    \begin{aligned}
    \mathcal{L}_{\textnormal{cos}}(G(w_a),x_a)=\max(d_\text{cos}(f_v(G(w_s)), f_a(x_a)) - \\
    d_\text{cos}(f_v(G(w_a)), f_a(x_a)) + 1, 0).
    \end{aligned}
    \label{loss:hinge}
\end{equation}
In the CLIP embedding space, the distance from the source latent vector $w_s$ is increased and closer to the audio-driven latent vector $w_a$. We minimize cosine distance between the latent representation of the manipulated image $G(w_a)$ and audio latent representation $f_a(x_a)$. $f_a$ and $f_v$ denote the pre-trained audio and visual encoders.

\begin{figure*}[t!]
  \centering
  \includegraphics[width=\textwidth]{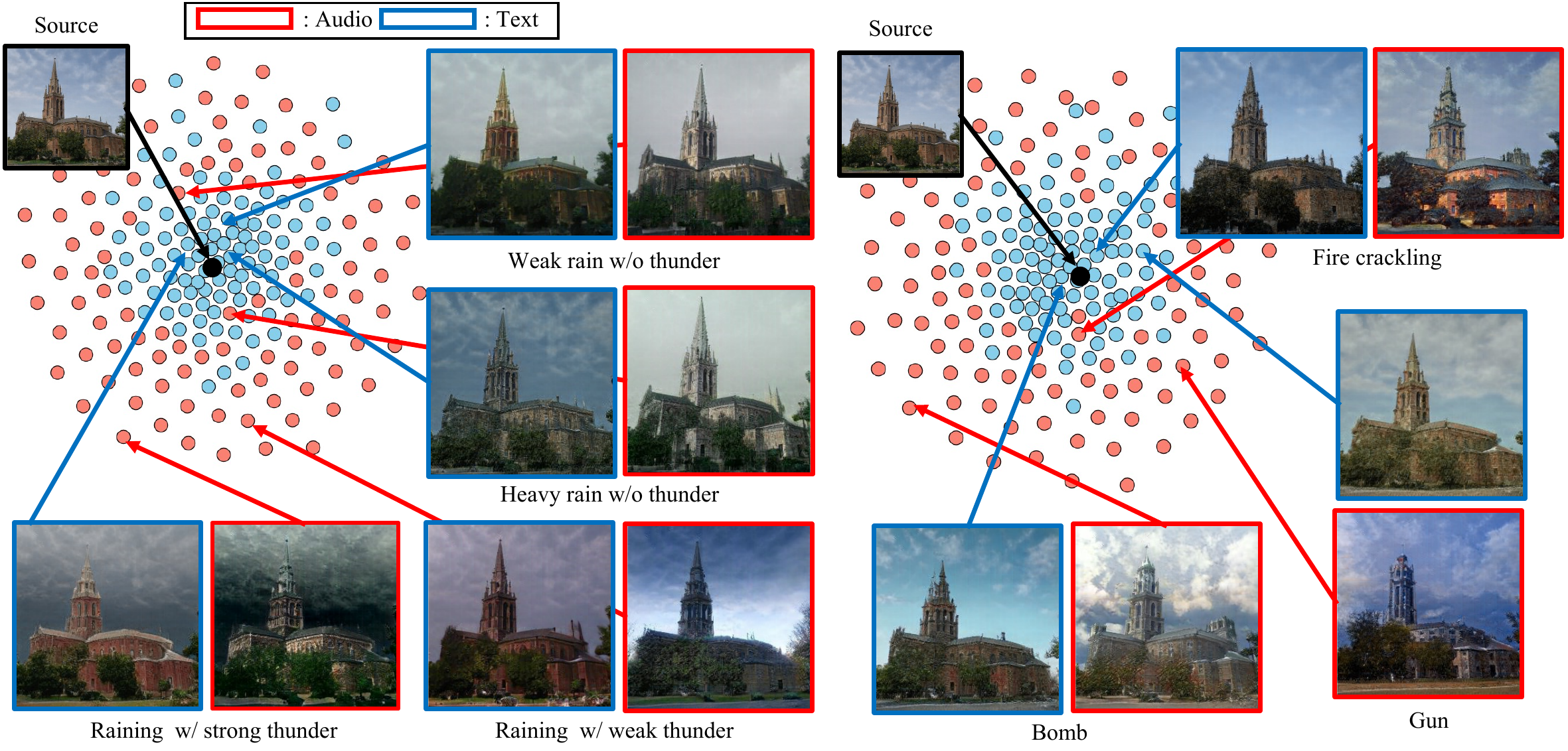}
  \caption{Comparison of manipulation results between our approach~(red) and the existing text-driven manipulation approach, StyleCLIP~\citep{Patashnik_2021_ICCV}~(blue). Unlike the text-based approach, we can produce more varied manipulation results, depending on the intensity of the sound. These are the results of image manipulation with weak rain, heavy rain, raining with weak thunder, raining with strong thunder, fire crackling, gun sounds, and bomb sounds from the left column to the right. We also report audio and text-guided manipulation directions with t-SNE visualization~\citep{van2008visualizing} (red: audio, blue: text).}
  \label{fig:difference}
\end{figure*}

\subsubsection{Adaptive Layer Masking} 
Generally, $L_2$ regularization effectively keeps the image generated from the moved latent code from being different from the original~\citep{Patashnik_2021_ICCV}. In the StyleGAN latent space, we maintain the image similarity by minimizing the $l_2$- distance between $w_a$ and $w_s$ as follows: 
\begin{equation}
    \begin{aligned}
        \mathcal{L}_{\textnormal{reg}}=||w_a - w_s||_2.
    \end{aligned}
\end{equation}
However, StyleGAN's latent code has different properties at each layer, so different weights should be applied to each layer if the user-provided attribute changes. We control style changes with adaptive layer masking. Layerwise masking keeps compact content information within style latent code as follows:
\begin{equation}
    \begin{aligned}
        \mathcal{L}_{\textnormal{reg}}=\cfrac{1}{L}\sum_{i=1}^{L}{g}_i \cdot|| (w_i^a - w_i^s)||_2.
    \end{aligned}
\end{equation}
In StyleGAN2~\citep{karras2020analyzing}, the latent code is represented as $ w \in \mathcal{W}^{L \times D}$, where $L$ is the number of layers and $D$ is the latent code's dimension. We declare a parameter vector $g \in \mathbb{R}^L$.  $g$ is a trainable vector that masks adaptively the specific style layer. 
$g$ ranks which of the $L$ layers is regulated by the softmax function. In the latent optimization step, $g$ and $w$ are multiplied according to the number of layers. $g$ is iteratively updated, which adaptively manipulates the latent code.

\subsubsection{Identity Loss}
The similarity to the input image is also controlled by the identity loss function $\mathcal{L}_{\textnormal{ID}}$, which is defined as:
\begin{equation}
    \mathcal{L}_{\text{ID}}(w_a) = 1 - \langle R(G(w_s)), R(G(w_a)) \rangle.
\end{equation}
where $R$ is the pre-trained ArcFace~\citep{deng2019arcface} model for face recognition. Thus, this loss function minimizes the cosine distance $\langle R(G(w_s)), R(G(w_a)) \rangle$ between its arguments in the latent space of the ArcFace network. This allows for manipulating facial expressions without changing the personal identity. Note that we disable the identity loss by setting $\lambda_{\textnormal{ID}}=0$ for all other non-human image manipulations.

\section{Experiments}

\subsection{Implementation and Training Details}
To learn the multi-modal joint embedding space, we use the pre-trained CLIP~\citep{radford2learning}-based image and text encoders, and we train a Swin transformer~\citep{liu2021swin}-based audio encoder, which outputs a latent vector of the length (i.e., 512) that is the same as the output of the image and text encoders. Note that we convert audio inputs into Mel-spectrogram acoustic features by following a common practice in various audio processing tasks. For the image manipulation step, we leverage the pre-trained StyleGAN2~\citep{karras2020analyzing} generator. Note that we adjust the size of the latent code depending on the resolution of the output image, i.e., $18 \times 512$ latent code for images of size $1024 \times 1024$ and $14 \times 512$ for $256 \times 256$. To implement such a generator, we utilize the official PyTorch implementation from StyleGAN2-ADA\footnote{https://github.com/NVlabs/stylegan2-ada-pytorch}.

We train our model for 50 epochs using the Adam optimizer with the cosine cyclic learning rate scheduler~\citep{smith2017cyclical}. We set the learning rate to $10^{-3}$ with the momentum of $0.9$, and we set the weight decay to $10^{-4}$. The batch size is set to 384. For audio augmentation, we use SpecAugment~\citep{park19e_interspeech} with a frequency mask ratio of $0.15$ and a time masking ratio of $0.3$. For direct latent code optimization, $\lambda_\text{reg}$ and $\lambda_\text{ID}$ in Eq. (\ref{loss:man}) are set to $0.008$ and $0.004$ for the FFHQ dataset; and $0.002$ and $0$ for the LSUN dataset.

\begin{figure}[t!]
  \centering
  \includegraphics[width=\linewidth]{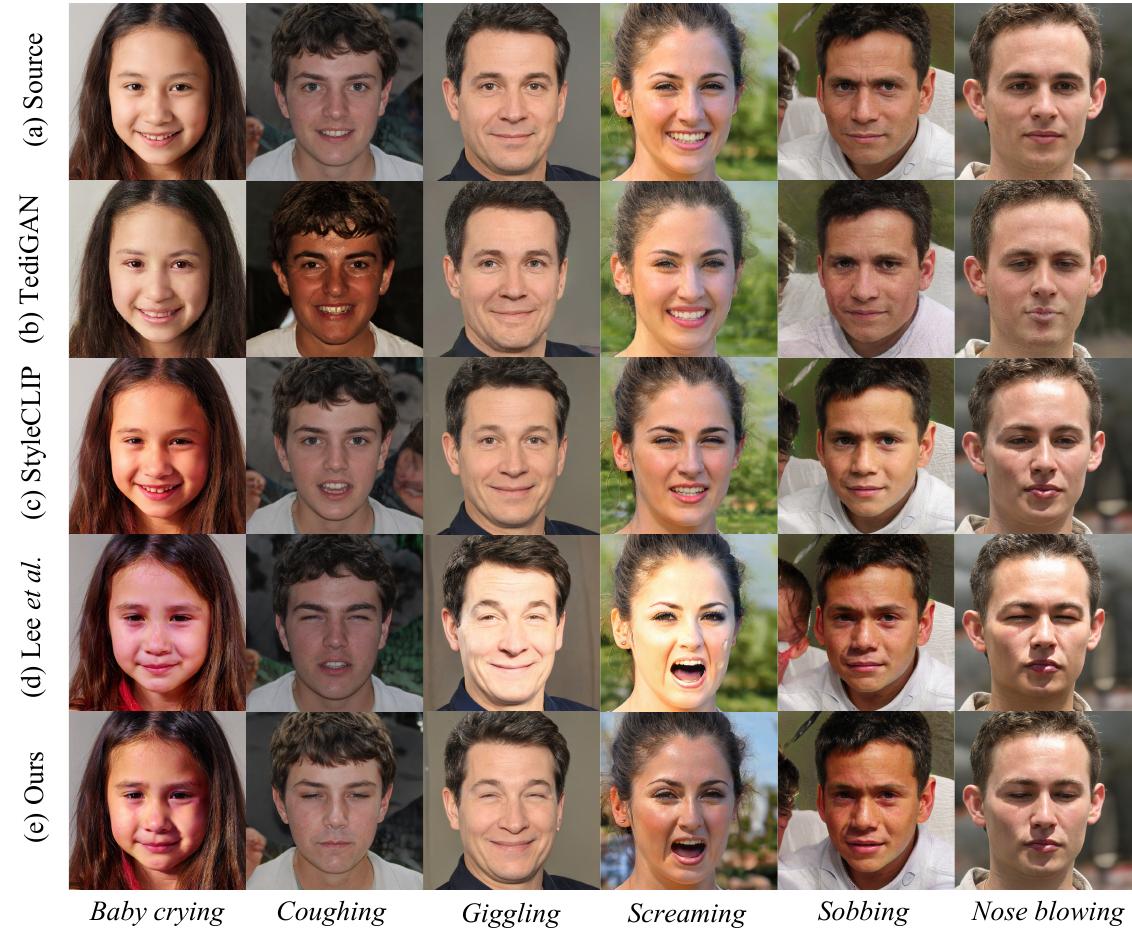}
  \caption{Given the (a) input image, we compare the face image manipulation results between (b-c) text-driven image manipulation approaches (i.e. TediGAN~\citep{xia2021tedigan} and StyleCLIP~\citep{Patashnik_2021_ICCV}), (d) Lee~\textit{et al.}~(previous sound-guided image manipulation approach) and (e) ours. The attributes for driving such manipulations include baby crying, coughing, giggling, screaming, sobbing and nose blowing.}
  \label{fig:cvprfig2}
\end{figure}

\subsection{Datasets}
To learn a multi-modal embedding space, we use the audio-text pair from a publicly available video dataset called AudioSet~\citep{gemmeke2017audio}, which contains 2,084,320 sound clips (in 10-second segments) extracted from YouTube videos and provides annotations of 632 audio classes. Because some of the associated URLs on YouTube were missing, we use 17,153 audio clips (out of 20,371 in the balanced subsets) and 1,617,939 clips (out of 2,041,789 in the unbalanced subsets) for training.

We also utilize VGG-Sound~\citep{chen2020vggsound}, which similarly provides over 200k human-labeled video clips (in 10-second segments) and the 310 corresponding audio classes. We finally use 182,342 videos for training due to missing clips. Note that a pair of images from the middle frame of each video clip and its corresponding class label is used as an image-text input pair for training our model.

Further, we train our StyleGAN2-based generator with the following high-resolution image datasets: Flickr-Faces-HQ (FFHQ)~\citep{karras2019style}, Large-Scale Scene Understanding (LSUN) Challenge~\citep{yu2015lsun} dataset, WikiArt~\citep{saleh2016large}, or Landscapes High-Quality (LHQ)~\citep{Skorokhodov_2021_ICCV}. Flickr-Faces-HQ (FFHQ)~\citep{karras2019style} contains the 70,000 high-quality human face images with a resolution of $1024 \times 1024$, and the Large-scale Scene Understanding (LSUN) Challenge~\citep{yu2015lsun} contains church images with a resolution of $256 \times 256$ and car images with a resolution of $384 \times 512$. WikiArt~\citep{saleh2016large} contains images of the paintings with a resolution of 1024~$\times$~1024 drawn by 195 different artists. Lastly, the Landscapes High-Quality~(LHQ)~\citep{Skorokhodov_2021_ICCV} contains nature landscape images of the resolution $256 \times 256$.

\subsection{Qualitative Analysis}
\subsubsection{Comparison Against Text-Guided Manipulation} 
We first compare our sound-guided image manipulation results with existing text-guided manipulation approaches, including TediGAN~\citep{xia2021tedigan} and StyleCLIP~\citep{Patashnik_2021_ICCV}. We also compare our approach with the sound-guided image manipulation approach from our prior work~\citep{lee2022sound}. As can be observed in Fig.~\ref{fig:cvprfig2}, image manipulation driven by sounds produces better manipulation quality (compare (b) and (c) vs. (d) and (e)). Furthermore, unlike text-guided methods, the audio-guided approach achieves natural image style transfer and is capable of representing multiple labels. For example, TediGAN emphasizes crying, whereas StyleCLIP focuses on the baby when the text ``baby crying'' is given. On the contrary, our proposed method can simultaneously handle ``baby'' and ``crying''. Moreover, our proposed method produces more visually plausible manipulation results than does our prior work~\citep{lee2022sound}, which often produced color shifts (e.g., skin tones).

\subsubsection{Audio Intensity-Aware Image Manipulation} 
We demonstrate that each audio sample has its own context, which makes the guidance in StyleGAN latent space richer than the text-based image manipulation method~(see Fig.~\ref{fig:difference}). Specifically, our sound representations provide guidance that correlates with the audio intensity in the StyleGAN latent space. If the magnitude of \textit{Thunder} is altered or if a specific attribute like \textit{Rain} is added to the audio, the style variation for the image is better correlated to the audio intensity than text-guided image manipulation. Likewise,~\textit{Fire crackling, Gunfire}, and~\textit{Bomb} sounds contain similar audio characteristics where something explodes, but our image manipulation results show the changes from the audio intensity. In particular, we would like to emphasize that no such explicit representation, such as \textit{Weak thunder} in the text descriptions, is used for pre-training the multi-modal embedding space.

We visualize the direction vector with t-SNE~\citep{van2008visualizing}~(see Fig.~\ref{fig:difference}). By subtracting the vectors of the latent code guided by each modality and the source latent code, we show the distribution of the direction of manipulation. 
We select the attributes in VGG-Sound~\citep{chen2020vggsound} and randomly manipulate the audio and text prompts.
Although we randomly sample the audio and text having the similar semantics, the sound-guided latent code shows a more significant transition than the text-guided latent code. 

\begin{figure}[t!]
  \centering
  \includegraphics[width=\linewidth]{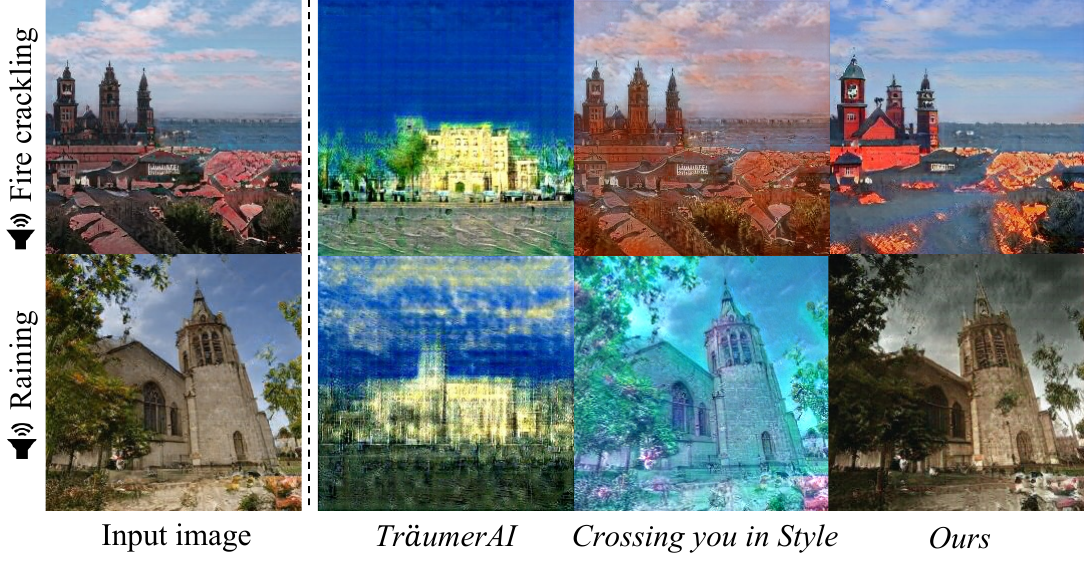}
  \caption{Comparison of sound-guided manipulation results. Given fire crackling (top) and raining (bottom) audio inputs, we manipulate the input image with Tr\"{a}umerAI~\citep{jeong2021tr}, Crossing you in style~\citep{lee2020crossing}, and our method.}
  \label{fig:cvprfig5}
\end{figure}

\begin{figure*}[t!]
  \centering
  \includegraphics[width=\textwidth]{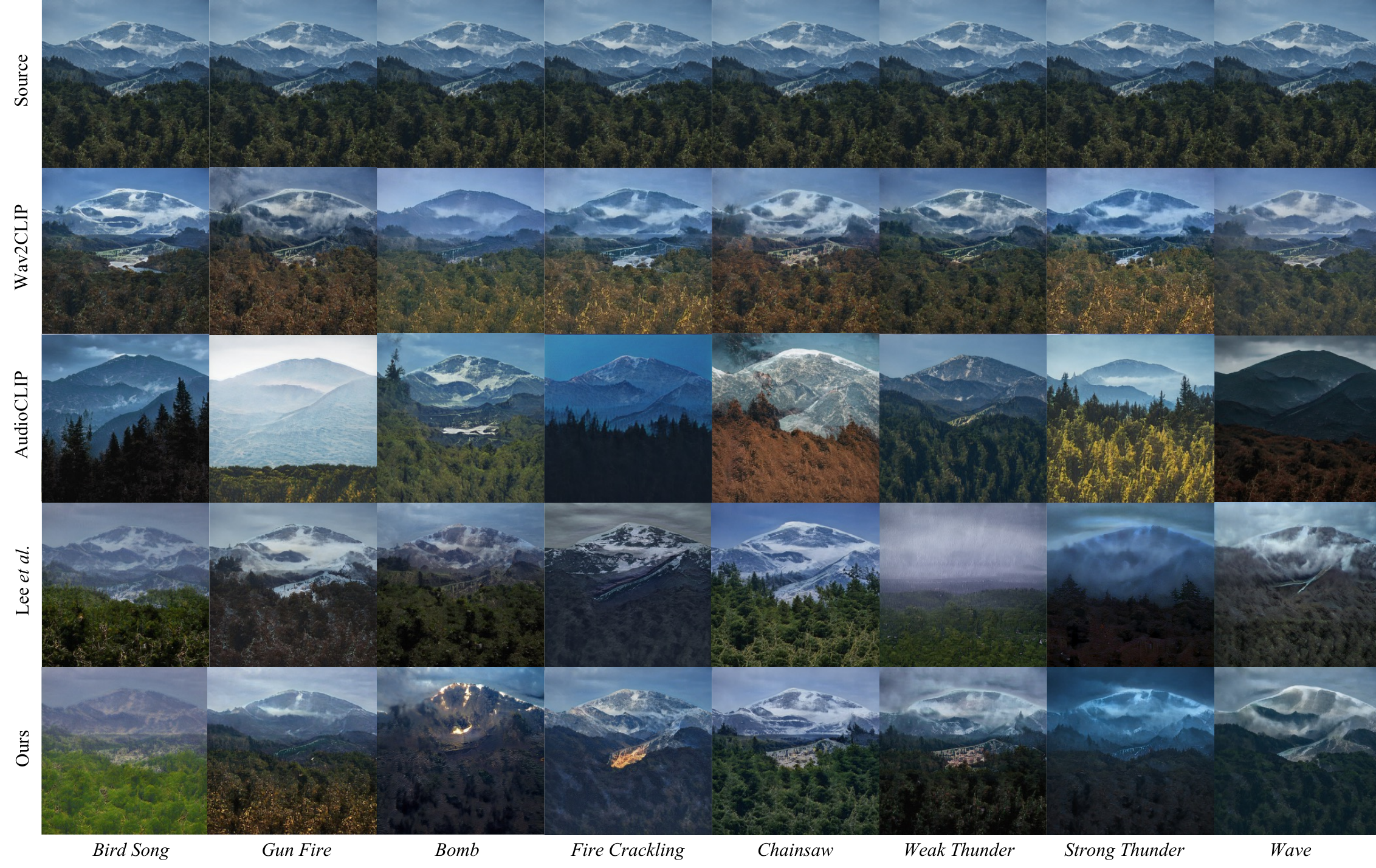}
  \caption{Image manipulation according to sound embedding space. We manipulate the input image with Wav2CLIP~\citep{wu2021wav2clip}, AudioCLIP~\citep{guzhov2021audioclip}, Lee~\textit{et al.}~\citep{lee2022sound} and our method. The sound attributes for image manipulation include~\textit{Bird song},~\textit{Gun fire},~\textit{Bomb},~\textit{Fire crackling},~\textit{Chainsaw},~\textit{Weak Thunder},~\textit{Strong Thunder}, and~\textit{Wave} sounds from left to right. Our model produces the most realistic sound-guided image manipulation results.}
  \label{fig:vsaudioclip}
\end{figure*}

\subsubsection{Comparison with Existing State-of-the-Art Sound-Based Style Transfer Models}
We compare our sound-guided image manipulation model with existing sound-based style-transfer models, including Tr\"{a}umerAI~\citep{jeong2021tr} and Crossing you in Style~\citep{lee2020crossing}. Fig.~\ref{fig:cvprfig5} compares image manipulation results in response to given audio inputs, e.g., audio of fire crackling and raining. We observe that our model produces a better quality of manipulated images. Existing models fail to capture semantic cues from the given audio input~(See 2\textsuperscript{nd} and 3\textsuperscript{rd} columns). In contrast, our model manipulates images to be consistent with the semantic cues from the input sound.

\subsubsection{Comparison with Existing Multi-modal Embedders}
Similar to our image-text-sound joint embedder, several works~\citep{wu2021wav2clip, guzhov2021audioclip, lee2022sound} have attempted to extend the CLIP model to jointly represent audio, text, and images. Here, we also compare those works with our learned embedding space to analyze their effectiveness for the image manipulation task. As shown in Fig.~\ref{fig:vsaudioclip}, we qualitatively compare image manipulation results based on different multi-modal joint embedding spaces learned by Wav2CLIP~\citep{wu2021wav2clip}, AudioCLIP~\citep{guzhov2021audioclip}, Lee~\textit{et al.}~\citep{lee2022sound}, and our joint embedder. Given different sound inputs (e.g., bird song, bomb, weak thunder, etc.), our learned embedding space generally produces the best results compared with the other embedders~(see 2nd, 3rd, and 4th rows vs. 5th row). This may confirm that our multi-modal joint embedding space is more effective for the image manipulation task than are existing approaches.

\begin{figure}[t!]
  \centering
  \includegraphics[width=\textwidth]{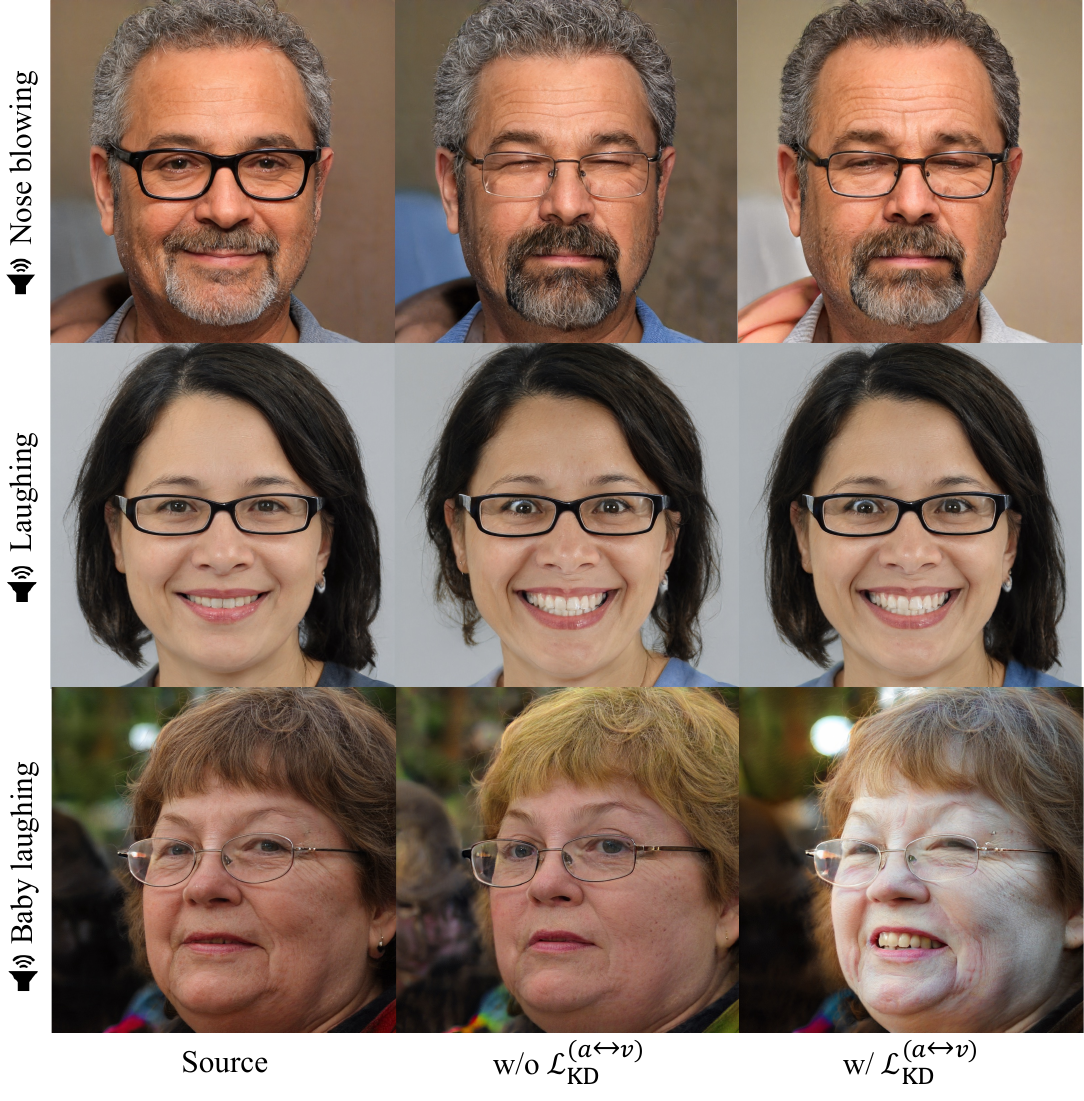}
  \caption{Ablation study of weakly paired contrastive learning for audio-visual joint embedding space. The leftmost is the input source, the middle is the manipulated result without KL divergence loss and the rightmost is the manipulation result after applying the KL divergence loss. In the middle, fine features such as hair loss change, whereas our method with KL divergence loss preserves the characteristics of the input source.}
  \label{fig:ablation_kd}
\end{figure}

\subsubsection{Effect of Audio-Visual Weakly Paired Contrastive Learning}
Weakly paired audio-visual contrastive learning makes the manipulation results robust. As demonstrated in Fig.~\ref{fig:ablation_kd}, we demonstrate that using weak supervision is helpful for reasonable sound-guided image manipulation. For example, manipulated images maintain the audio-irrelevant attributes like glasses, hair styles and beards. Moreover, it is more robust for manipulating target attributes. It shows the manipulation results that satisfy the keywords "baby" and "laughing" at the same time.

\subsubsection{Effect of Adaptive Layer Masking}
\begin{figure}[t!]
  \centering
  \includegraphics[width=\linewidth]{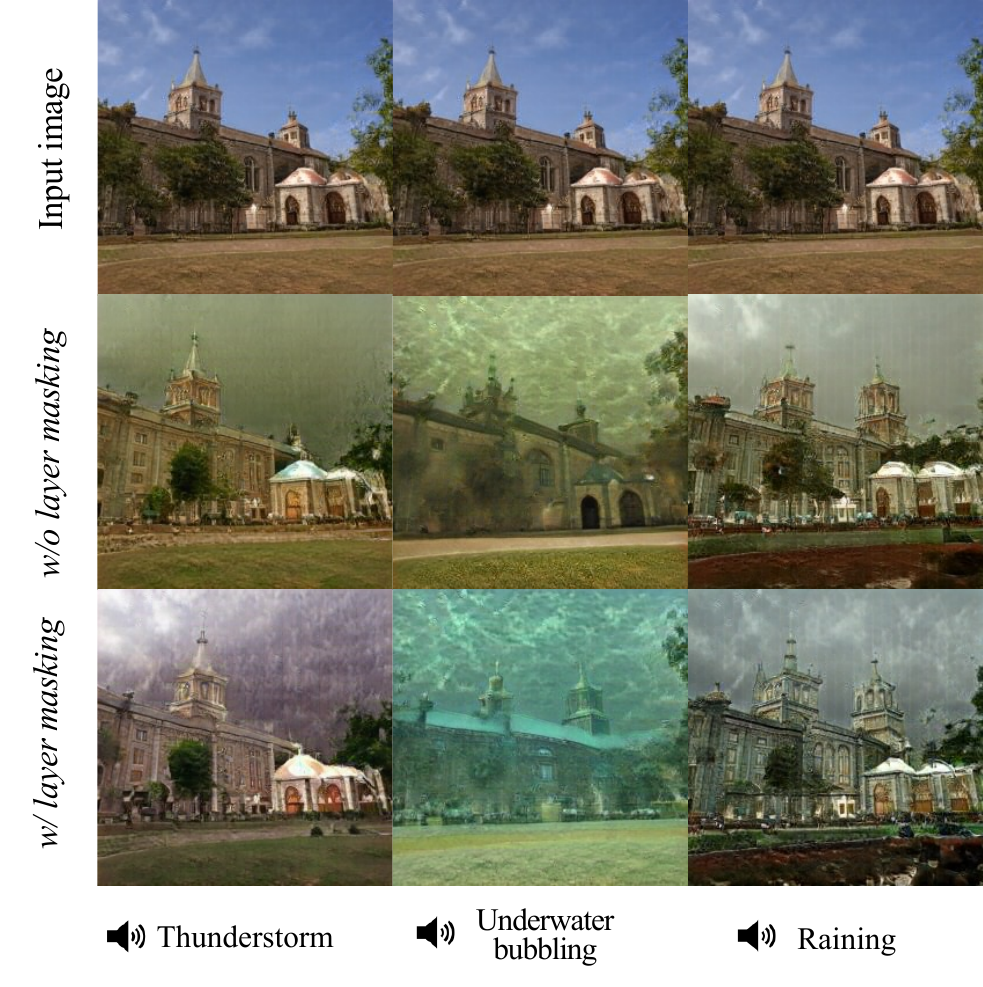}
  \caption{Ablation study of adaptive layer masking. The first row shows the input image, the second row shows the manipulation results without adaptive layer masking, and the third row shows the sound-guided image manipulation results after applying the adaptive layer masking. }
  \label{fig:cvprfig6}
\end{figure}

In StyleGAN~\citep{jeong2021tr}, it is necessary to adaptively regularize the style layer, because each layer of latent code has different style attributes. Each layer of latent code multiplies the trainable parameter that controls the diversity during regularization. The ablation study shows a qualitative comparison of the mechanism for applying adaptive layer masking to the style layer, as illustrated in Fig.~\ref{fig:cvprfig6}. The adaptive masking rectifies the direction by changing the latent code based on the semantic cue. When the gate function is applied, sound-guided image manipulation is semantically reasonable. For example, a thunderstorm is a mixture of thunder and rain sound. Although thunder and lightning are not seen in the second row, lightning and rain appear in the last row.

\begin{table}[t!]
    \caption{Comparison of the quality of audio representations between ours and alternatives. We report the classification accuracy (top-1 in \%) of a linear classifier on the ESC-50~\citep{piczak2015esc} and the Urban sound 8k~\citep{Salamon:UrbanSound:ACMMM:14} datasets as well as their zero-shot inference results. {\it{Abbr.}} \it{S}: supervised setting.}
    \label{zero}
    \centering
    \resizebox{\linewidth}{!}{
        \begin{tabular}{@{}lcccc@{}}
            \toprule
            \multirow{2}{*}{Model} & \multirow{2}{*}{\it{S}} & \multirow{2}{*}{Zero-shot} & \multicolumn{2}{c}{Dataset} \\ \cmidrule{4-5}
            & & & ESC-50~($\uparrow$) & Urban sound 8k~($\uparrow$) \\ 
            \midrule
            ResNet50~\citep{hershey2017cnn} & \checkmark & - & 66.8 \% & 71.3 \% \\ 
            Lee~\textit{et al.}~\citep{lee2022sound}~(LR) & - & - & 72.2 \% & 66.8 \% \\
            Ours~(LR) & - & - & \textbf{89.5 \%} & \textbf{72.4 \%} \\
            \midrule
            Wav2clip~\citep{wu2021wav2clip} & - & \checkmark & 41.4 \% & 40.4 \%  \\
            AudioCLIP~\citep{guzhov2021audioclip} & - & \checkmark & 69.4 \% & \textbf{68.8} \% \\
            Lee~\textit{et al.}~\citep{lee2022sound} & - & \checkmark & 57.8 \% & 45.7 \%  \\
            Ours w/o $\mathcal{L}_\text{KL}^{(a \leftrightarrow \tilde{v})}$ & - & \checkmark & 63.6 \% & 53.2 \% \\
            Ours & - & \checkmark & \textbf{74.4 \%} & 58.6 \%  \\
            \bottomrule
        \end{tabular}
    }
\end{table}

\subsection{Quantitative Analysis}
\subsubsection{Zero-Shot Transfer} 
We compare our model to the supervised method and the existing zero-shot audio classification method~(see Table~\ref{zero}). First, we compare audio embeddings trained by the supervised methods such as logistic regression and ResNet50~\citep{hershey2017cnn} supervised by random initialization of weights as a baseline model. Our audio encoder shows better classification performance than the baseline. Secondly, we compare the zero-shot audio classification accuracy with Wav2CLIP~\citep{wu2021wav2clip} and AudioCLIP~\citep{guzhov2021audioclip}. Our proposed loss function learns three modalities in the CLIP embedding space and obtains a richer audio representation through contrastive loss. However, Wav2CLIP only understands the relationship between the audio and the visual context. Furthermore, Wav2CLIP learns from an audio-visual having a dataset with a relatively small data scale to distillate the CLIP representation to the audio encoder~(the VGG-Sound dataset: 200K). AudioCLIP uses a dataset scaled up from contrastive learning between modalities from the audio, image, and text given as a triplet to obtain a more powerful sound representation.
The AudioCLIP's audio encoder was pre-trained on a large-scale audio dataset, namely AudioSet~\citep{gemmeke2017audio}, which consists of over 2M sound clips.

As a result, Table~\ref{zero} also illustrates that the power of sound representation improves when the scale of positive audio-visual data is increased with weakly paired contrastive learning than Lee~\textit{et al.}~\citep{lee2022sound}'s work.
Moreover, our learned embedding space is more suitable for the image manipulation task.

\begin{table*}[t!]
  \caption{Cosine similarity comparison between text-guided latent code $w_t$ and sound-guided latent code $w_a$. Note that $w_t$ is the latent code from StyleCLIP~\citep{Patashnik_2021_ICCV}'s text-driven latent optimization. $w_s$ indicates the source latent code.}
  \label{tab:similarity}
  \centering
  \begin{adjustbox}{max width=\textwidth}
  \begin{tabular}{@{}cccccccccccccc@{}}
    \toprule
    \multicolumn{14}{c}{Attribute} \\\cmidrule{3-14}
    Latent code & Metric & \multirow{2}{*}{\parbox{1.8cm}{\centering Giggling}} & \multirow{2}{*}{\parbox{1.8cm}{\centering Sobbing}} & \multirow{2}{*}{\parbox{1.5cm}{\centering Nose blowing}} & \multirow{2}{*}{\parbox{1.8cm}{\centering Fire\\ crackling}} & \multirow{2}{*}{\parbox{1.8cm}{\centering Wind noise}} & \multirow{2}{*}{\parbox{1.8cm}{\centering Underwater\\ bubbling}} & \multirow{2}{*}{\parbox{1.8cm}{\centering Explosion}} & \multirow{2}{*}{\parbox{1.8cm}{\centering Thunderstorm}} & \multirow{2}{*}{\parbox{1.8cm}{\centering Gun}} &  \multirow{2}{*}{\parbox{1.8cm}{\centering Raining}}  & \multirow{2}{*}{\parbox{1.8cm}{\centering Bird song}} & \multirow{2}{*}{\parbox{1.8cm}{\centering Average}} \\ \\
    \midrule
    ($w_s$, $w_a$) & Mean~($\downarrow$) & 0.99315  & 0.99354 & 0.99201 & 0.97400  & 0.97333 & 0.97409 & 0.97230 & 0.97014 & 0.97645 & 0.97588 & 0.97613 & 0.97918 \\
                   & Std~($\uparrow$) & 0.00154 & 0.00168 & 0.00267 & 0.00868 & 0.00881 & 0.00929  & 0.01042 & 0.00984 & 0.00755 & 0.00898 & 0.00837 & 0.00708 \\
    \midrule
    ($w_s$, $w_t$) & Mean~($\downarrow$) & 0.99866  & 0.99849 & 0.99779 & 0.99260  & 0.99095 & 0.99416 & 0.99437 & 0.99136 & 0.99208 & 0.99332 & 0.99238 & 0.99420 \\
                   & Std~($\uparrow$) & 0.00054 & 0.00034 & 0.00064 & 0.00301 & 0.00359 & 0.00246  & 0.00240 & 0.00369 & 0.00222 & 0.00259 & 0.00239 & 0.00217 \\
    \midrule
    ($w_a$, $w_t$) & Mean~($\downarrow$) & 0.99554  & 0.99510 & 0.99448 & 0.97519  & 0.97280 & 0.97082 & 0.9687 & 0.96926 & 0.97716 & 0.97093 & 0.97488 & 0.97862 \\
                   & Std~($\uparrow$) & 0.00158 & 0.00170 & 0.00213 & 0.00937 & 0.01026 & 0.01035  & 0.01160 & 0.01045 &  0.00813 & 0.01023 & 0.00912 & 0.00772 \\
    \bottomrule
  \end{tabular}
  \end{adjustbox}
\end{table*}

\begin{figure*}[t!]
  \centering
  \includegraphics[width=\textwidth]{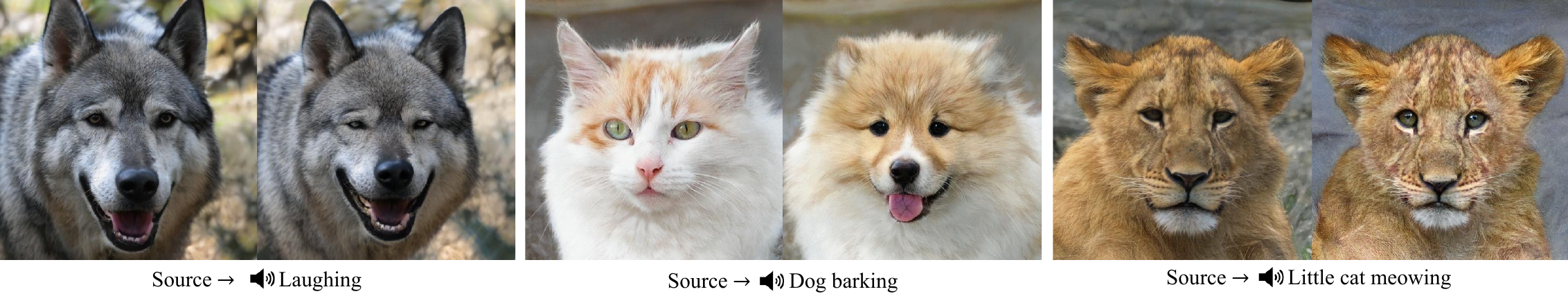}
  \caption{Examples of sound-guided animal face image manipulation results.}
  \label{fig:stylegan3}
\end{figure*}

\subsubsection{Semantic Accuracy of Manipulation} 
We quantitatively analyze the effectiveness of our proposed audio-driven image manipulation approach. First, we measure performance on the semantic-level classification task. Given the audio embeddings from our pre-trained audio encoder, we train a linear classifier to recognize eight semantic labels, including giggling, sobbing, nose-blowing, fire crackling, wind noise, underwater bubbling, explosion, thunderstorm, gun, raining, and bird song. We use StyleGAN2~\citep{karras2020analyzing} weights pre-trained from the FFHQ~\citep{karras2019style} dataset for the guiding with giggling, sobbing, and nose-blowing attributes to compare the semantic-level classification accuracy between text and audio. Also, for the guiding with fire crackling, wind noise, underwater bubbling, explosion, and thunderstorm attributes, the weights of StyleGAN2 pre-trained on the LSUN (church)~\citep{yu2015lsun} dataset are used. As shown in Fig.~\ref{fig:userstudy}~(a), we generally outperform the existing text-driven manipulation approach with better semantically-rich latent representation.

\subsubsection{Distribution of Manipulation Direction} 
We can see how much the latent code has changed by the cosine similarity between the source latent code and the manipulated latent code. We compare the cosine similarity between text-guided and sound-guided latent representations. We evaluate the mean and variance of the cosine similarity between $w_s$, a source latent code, $w_a$, an audio-driven latent code, and $w_t$, a text-driven latent code. The latent representations generally exhibit a high-level characteristic of the content~(see Table~\ref{tab:similarity}.). In the latent space of StyleGAN2, the sound-guided latent code moves more from the source latent code than does the text-guided latent code, and the image generated from the sound-guided latent code is more diverse and dramatic than for the text-guided method.

\subsection{Human Evaluations}
\begin{figure}[t!]
  \centering
  \includegraphics[width=\linewidth]{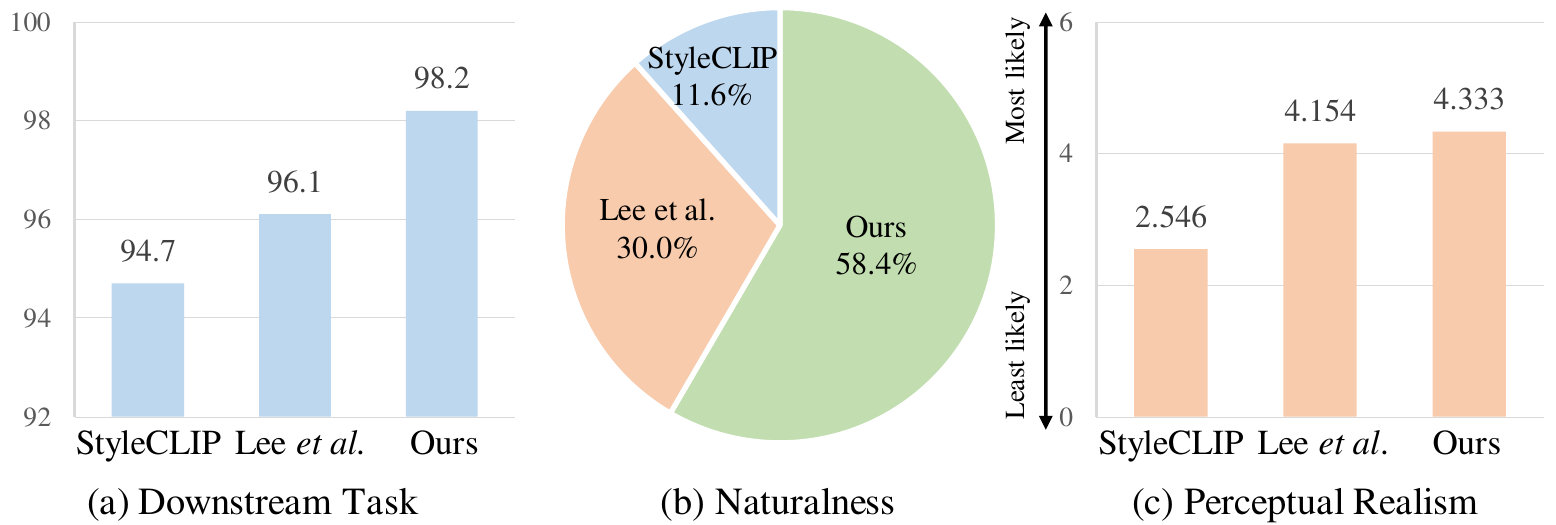}
  \caption{Quantitative evaluation and user study results. (a)~Downstream task evaluation to compare the quality of representations between our approach and text-driven manipulation approaches on the FFHQ~\citep{karras2019style} dataset. A linear classifier is trained to predict 11 semantic labels, e.g. giggling. Participants answered a questionnaires on (b)~Naturalness~(``Which image manipulation result better expresses the target attribute?'') and (c)~Perceptual Realism~(``On the scale from 1 to 5, how realistic are the images below?'').}
  \label{fig:cvprfig13}
  \label{fig:userstudy}
\end{figure}

We recruited fifty participants from Amazon Mechanical Turk~(AMT) to evaluate the effectiveness of our proposed method from a human perspective. We generate a set of samples of manipulated images from 12 different sound sources: e.g., laughing, sobbing, nose-blowing, wind noise, underwater bubbling, explosion, and thunderstorms. Source images are sampled from FFHQ~\citep{karras2019style}, LSUN~\citep{yu2015lsun}, and LHQ~\citep{Skorokhodov_2021_ICCV} datasets. In our human study, participants need to complete the following two questionnaires: (i) Naturalness-~\textit{Which image manipulation result better expresses the target attribute?} and (ii) Perceptual Realism-~\textit{How realistic are the images below?} Note that we use the Likert scale ranging from 1~(low realistic) to 5~(high realistic) for perceptual realism, and we compare our manipulation results with StyleCLIP~\citep{Patashnik_2021_ICCV} and work by Lee \textit{et al.}~\citep{lee2022sound}. As we summarized in Fig.~\ref{fig:userstudy} (b) and (c), our model clearly outperforms other state-of-the-art approaches (StyleCLIP and Lee~\textit{et al.}) in both human evaluations. A large portion of participants~($58.4$\%) judge our generated images to be better manipulated than the images of other approaches, and our method obtains a score of 4.333 on average in terms of perceptual realism. This may confirm our model generates more perceptually realistic images than the other models.

\begin{figure}[t!]
  \centering
  \includegraphics[width=\linewidth]{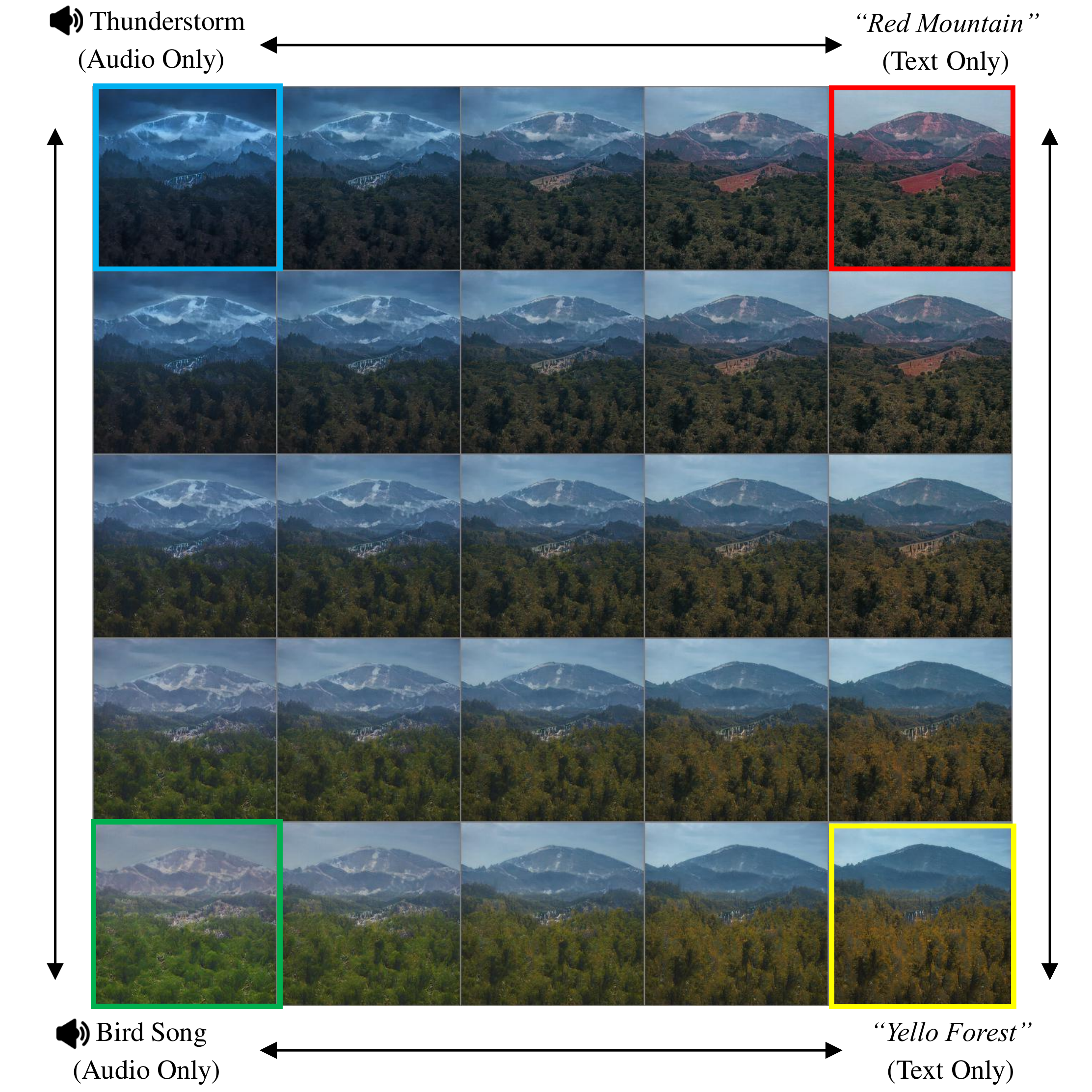}
  \caption{Manipulation results of interpolation between interpolation between text and sound-guided StyleGAN latent vectors. Even if the source latent vector is guided by a different modality, we can control the style changes linearly.}
  \label{fig:interpolation}
\end{figure}

\begin{figure}[t!]
  \centering
  \includegraphics[width=\linewidth]{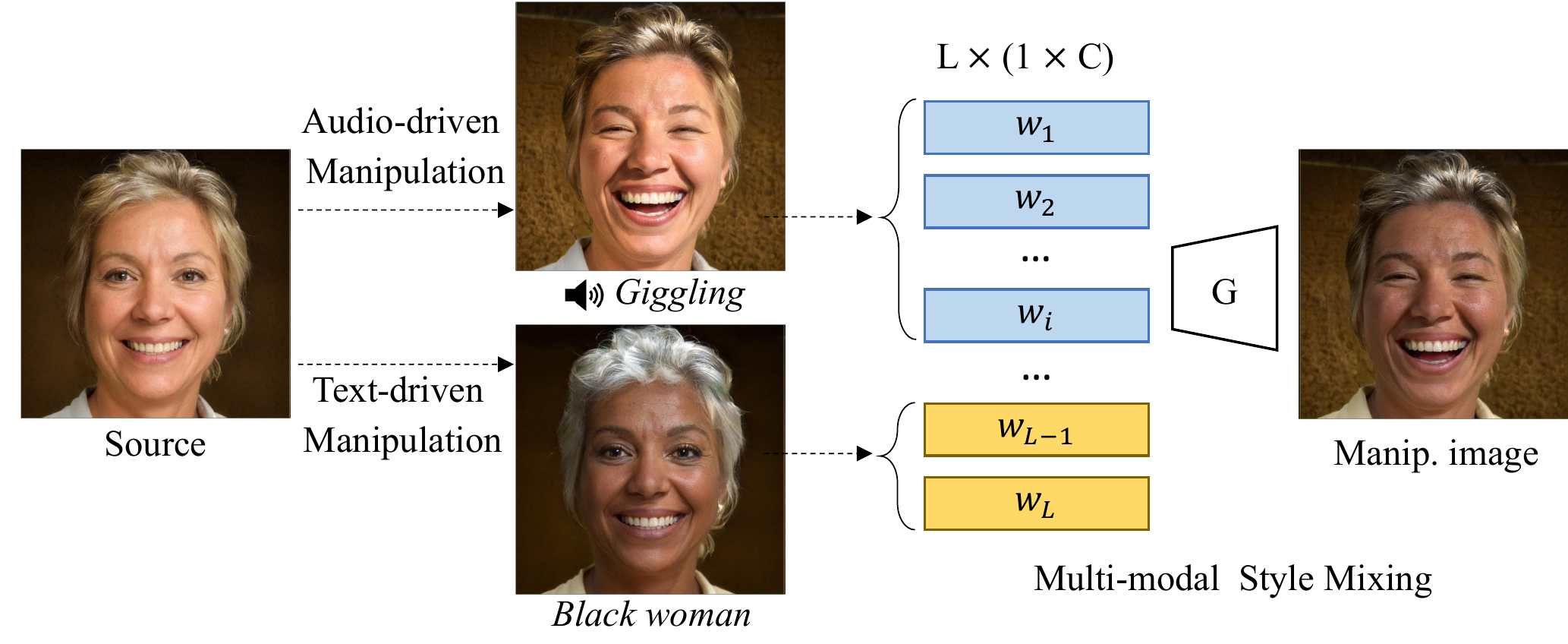}
  \caption{Overview of image style mixing jointly with the audio (\textit{people giggling}) and text input (\textit{black woman}).}
  \label{fig:stylemixing}
\end{figure}

\subsection{Sound and Text-Guided Style Mixing} 
Creating new imaginary visual content by combining text and audio information is promising.
Our method ensures that audio, text, and image share the same embedding space. We interpolated text and sound-guided latent code to demonstrate that multi-modal embedding lies in the same latent space~(see Fig.~\ref{fig:interpolation}). We obtain the interpolated latent style vector as follows:
\begin{equation}
    \begin{aligned}
        w=(1-\alpha)\cdot w_a + \alpha \cdot w_t.
    \end{aligned}
\end{equation}
$w$ denotes the interpolated latent vector which is the weighted sum of $w_t$, the text-guided latent vector, and $w_a$ denotes the sound-guided latent vector. The results generated from the interpolated latent vector contains the continuous intermediate meaning of the two modalities.

We propose a multi-modal manipulation of audio and text based on the style mixing technique of StyleGAN. Different layers of $w$ latent code in StyleGAN represent different properties. We perform multi-modal style mixing by selecting a specific layer of latent code and a mixing style that includes audio and text. Selecting a particular layer of each latent code guided by audio and text manipulates the image using the properties of audio and text~(see Fig.~\ref{fig:stylemixing}). 

We find that the sound source can effectively manipulate aspects of facial emotion  such as ``giggling" on the face and text information controls the background color of the target image. For the style-mixing details, we follow TediGAN's StyleGAN layerwise analysis~\citep{xia2021tedigan}. In the 18 $\times$ 512 latent code, the style-mixing technique selects the 1st to 9\textsuperscript{th} layers of the sound-guided latent code and the 10\textsuperscript{th} to 18\textsuperscript{th} layers of the text-guided latent code to mix the dynamic characteristics of sound and the human properties of text.

\section{Applications}
\subsection{Sound-Guided Animal Face Image Manipulation}
Sound-guided Animal Face Image Manipulation
We further apply our method to manipulate the images of animal faces. We pre-train the StyleGAN3~\citep{karras2021alias} generator with the existing Animal FacesHQ (AFHQ) v2~\citep{choi2020stargan} dataset, which contains over 15k images of multiple species~(cat, dog, wildlife) and various breeds ($\geq$ eight) per species. As shown in Fig.~\ref{fig:stylegan3}, our model successfully manipulates animal faces with guidance by a sound input~(e.g., laughing, dog barking, and little cat meowing).

\subsection{Sound-Guided Artistic Painting Manipulation} 
\begin{figure}[t!]
  \centering
  \includegraphics[width=\linewidth]{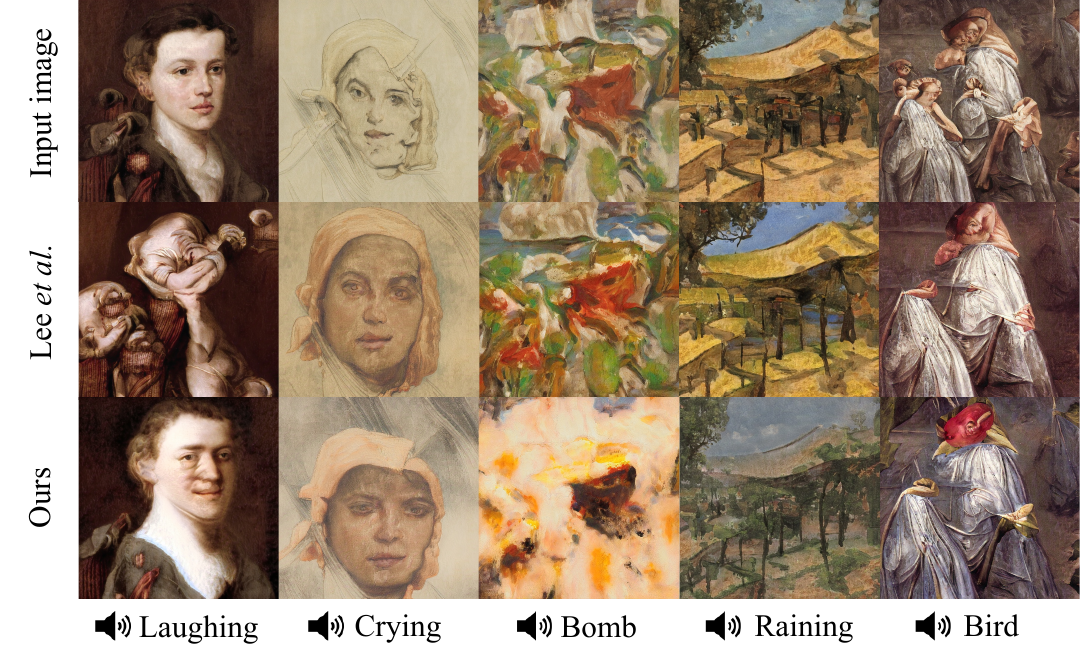}
  \caption{Examples of sound-guided artistic painting manipulation results.}
  \label{fig:paintings}
\end{figure}

We further apply our approach for sound-guided artistic painting manipulation. We train our StyleGAN2~\citep{karras2020analyzing}-based generator with a fine-art paintings dataset, called WikiArt~\citep{saleh2016large}, which contains over 50k images of paintings from 195 different artists. We observe in Fig.~\ref{fig:paintings} that our model successfully produces manipulated images for art paintings guided by audio inputs (e.g., laughing, crying, bomb, raining, or bird). For example, given a laughing sound as a semantic cue, a portrait painting of a woman is manipulated (see 1st column), but the existing approach by Lee~\textit{et al.} does not. Our human study (done in the same setting as our user study) also confirms that 61.2\% of 50 participants judge that our results are better manipulated than those by Lee~\textit{et al.}~(26.7\%) and text-based approach~(12.1\%).

\subsection{Music Style Transfer} 
\begin{figure}[t!]
  \centering
  \includegraphics[width=\linewidth]{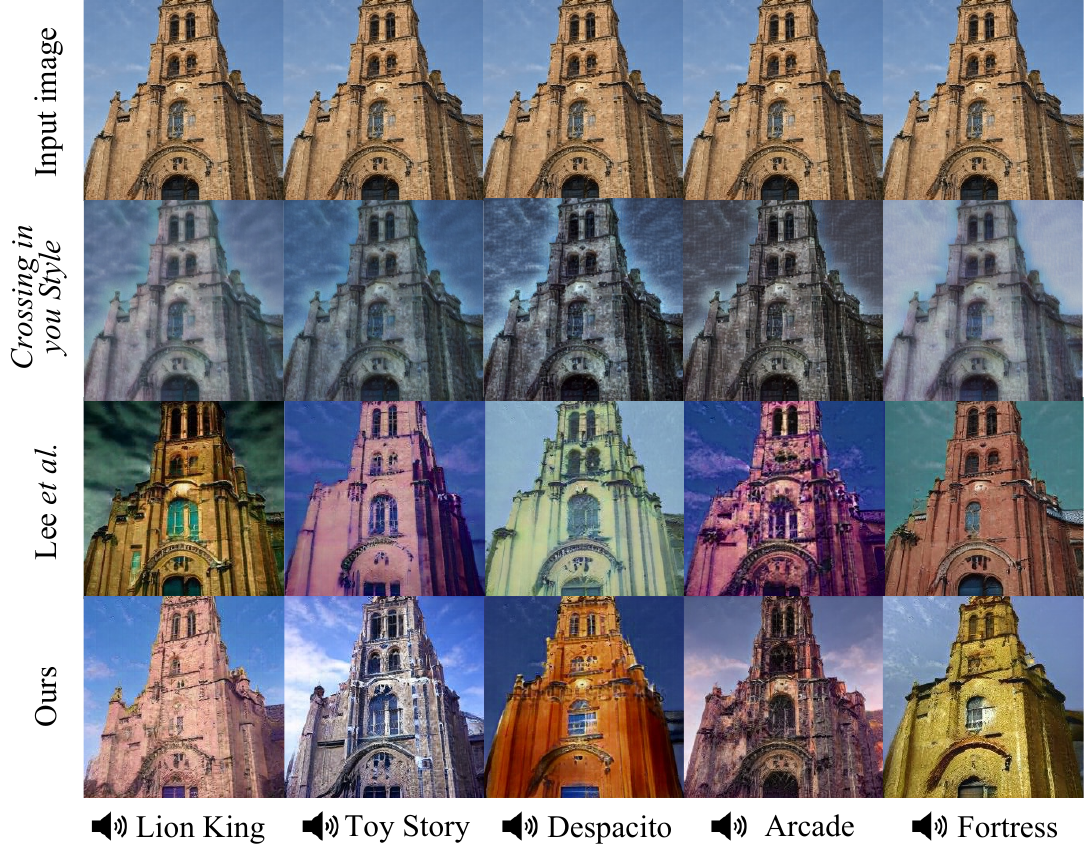}
  \caption{Examples of music style transfer using our method. Given 10 seconds of music, we manipulate the input image using Crossing you in style~\citep{lee2020crossing}, Lee~\textit{et al.}~\citep{lee2022sound} and our method.}
  \label{fig:musics}
\end{figure}

The objective of music style transfer's objective is that the source latent code is close to the keywords of the music, so the mood of the music appears in the image. For instance, \textit{Funny} music manipulates the image with a fairy-tale style, whereas \textit{Latin} music manipulates the image with red-colored theme that reflects the characteristic of \textit{passion}. Lee~\textit{et al.}~\citep{lee2022sound} demonstrate that CLIP-based sound representation can potentially reflect the music's mood into the image style. In this study, we outperform previous works on visual style transfer for music. We collect well-known 10-second segments of well-known music and compare the image manipulation results~(\textit{The Lion King} OST, \textit{The Toy Story} OST,~\textit{Despacito} title song, game sound effects,~\textit{Fortress} game song) for evaluation. Fig.~\ref{fig:musics} illustrates the results of image style transfer using various music genres. Visual information collected from text descriptions related to music genres in the AudioSet results in robust image manipulation for the music the model encounters for the first time.

\section{Discussion}
\subsection{Limitations} 
Our method performs sound-guided image manipulation for various domains and synthesizes realistic manipulation results. However, our model requires a trained StyleGAN latent space, which increases the learning cost in real applications and limits the domain of visual content. In order to manipulate images in the open domains, it is necessary to obtain a prior that directly generates the image from the latent representation of the sound, rather than providing guidance to the StyleGAN latent space as a CLIP-based sound embedding space. This will be studied as future work.

\subsection{Societal Impacts} 
It can have consequences for misuse in creating fake news or propaganda. Moreover, the method of sound-guided image manipulation is based on CLIP's knowledge and may reflect social prejudice. In the CLIP paper, the author says ``CLIP is trained on text paired with images on the internet. These image-text pairs are unfiltered and uncurated and result in CLIP models learning many social biases.''. Therefore, there is a possibility that certain social biases may appear when people faces are manipulated with sounds related to a thief, prisoner, criminal, or suspicious people, such as guns and fighting sounds. Also, when a human face is manipulated with a sound such as a vacuum cleaner or a women speaking, manipulation results with social prejudice like housekeepers may be generated.

\section{Conclusion}
We propose a method to manipulate images based on the semantic-level understanding from the given audio input. We take the user-provided audio input into the latent space of StyleGAN2~\citep{karras2020analyzing} and the CLIP~\citep{radford2learning} embedding space. In particular, the strategy of adding a semantic edge between audio-visual from text obtains a more robust sound representation for fine-features. Then, the latent code is aligned with the audio to enable meaningful image manipulation while reflecting the context from the audio. Our model produces responsive manipulations based on various audio inputs. We observe that an audio input can successfully provide a semantic cue to manipulate images accordingly. Our method of traversing multi-modal embedding space can be used in many applications with multi-modal contexts.

\section*{Acknowledgments}
\noindent This research was fully supported by Culture, Sports and Tourism R\&D Program through the Korea Creative Content Agency grant funded by the Ministry of Culture, Sports and Tourism in 2022~(Project Name: 4D Content Generation and Copyright Protection with Artificial Intelligence, Project Number: R2022020068).

\bibliographystyle{elsarticle-harv} 
\bibliography{main}





\end{document}